\documentclass[10pt,twocolumn,letterpaper]{article}

\usepackage{cvpr}              % To produce the CAMERA-READY version
\definecolor{cvprblue}{rgb}{0.21,0.49,0.74}
\usepackage[pagebackref,breaklinks,colorlinks,allcolors=cvprblue]{hyperref}
\usepackage{multicol,multirow,bm}
\usepackage{graphicx,subcaption}
\usepackage[table]{xcolor}
\usepackage{dblfloatfix}
\def\paperID{XXXXX} % *** Enter the Paper ID here
\def\confName{CVPR}
\def\confYear{2026}

\title{Rethinking MLLM Itself as a Segmenter with a Single Segmentation Token}

\author{
Anqi Zhang$^{1,2}$
\thanks{Equal contribution.}
\thanks{Work done during internship in University of Birmingham.}
\hspace{0.1cm}, 
Xiaokang Ji$^{1}$\footnotemark[1]\hspace{0.1cm},
Guangyu Gao$^{1}$\thanks{
Corresponding author, guangyugao@bit.edu.cn.
}\hspace{0.1cm},
Jianbo Jiao$^{2}$,
Chi Harold Liu$^{1}$,
Yunchao Wei$^{3,4}$\\
$^{1}$Beijing Institute of Technology\qquad 
$^{2}$University of Birmingham\qquad
$^{3}$Beijing Jiaotong University\\
$^{4}$Beijing Academy of Artificial Intelligence\
}

\begin{document}
\maketitle

\begin{abstract}

% Recent segmentation methods based on the Multi-modal Large Language Models~(MLLMs) enable reliable segmentation ability on a specific object, which activates more accurate spatial perception. 
Recent segmentation methods leveraging Multi-modal Large Language Models (MLLMs) have shown reliable object-level segmentation and enhanced spatial perception.
However, almost all previous methods predominantly rely on specialist mask decoders to interpret masks from generated segmentation-related embeddings and visual features, or incorporate multiple additional tokens to assist. 
% Our paper aims to investigate whether and how we can get rid of the external mask decoder and tokens while achieving competitive results. 
This paper aims to investigate whether and how we can unlock segmentation from MLLM itSELF with 1 segmentation Embedding~(SELF1E) while achieving competitive results, which eliminates the need for external decoders.
To this end, our approach targets the fundamental limitation of resolution reduction in pixel-shuffled image features from MLLMs.
First, we retain image features at their original uncompressed resolution, and refill them with residual features extracted from MLLM-processed compressed features, thereby improving feature precision. 
Subsequently, we integrate pixel-unshuffle operations on image features with and without LLM processing, respectively, to unleash the details of compressed features and amplify the residual features under uncompressed resolution, which further enhances the resolution of refilled features. 
Moreover, we redesign the attention mask with dual perception pathways, \textit{i.e.}, image-to-image and image-to-segmentation, enabling rich feature interaction between pixels and the segmentation token.
Comprehensive experiments across multiple segmentation tasks validate that SELF1E achieves performance competitive with specialist mask decoder-based methods, demonstrating the feasibility of decoder-free segmentation in MLLMs.
Project page: \url{https://github.com/ANDYZAQ/SELF1E}.
% we redesign the attention mask with image-to-image and image-to-segmentation perception, enabling adequate feature interaction among the pixels and segmentation token. 
% Extensive experiments show that our method achieves competitive results across several segmentation tasks compared to other methods with specialist mask decoders. 

\end{abstract}    
\section{Introduction}
\label{sec:intro}

\begin{figure}[!t]
    \centering
    \begin{subfigure}{0.25\textwidth}
        \centering
        \includegraphics[width=0.75\linewidth]{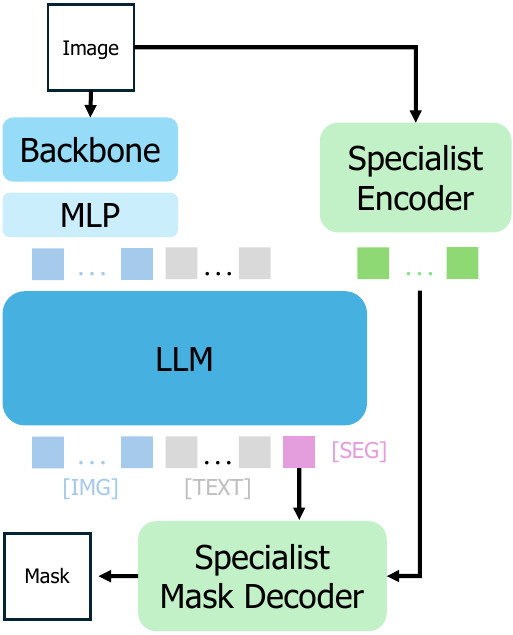} 
        \caption{MLLM with specialist encoder \& decoder.}
        \label{fig:introcom_spe}
    \end{subfigure}
    \hfill 
    \begin{subfigure}{0.2\textwidth}
        \centering
        \includegraphics[width=0.75\linewidth]{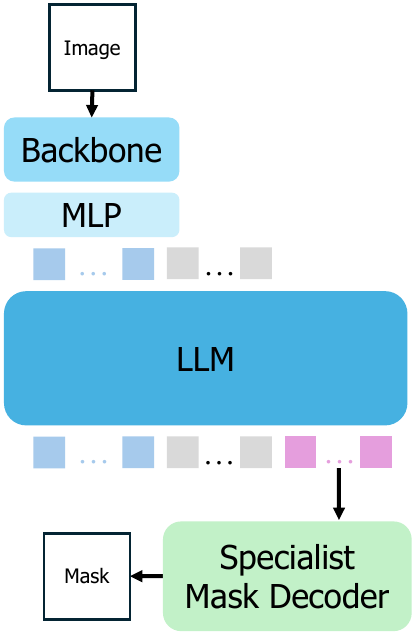}
        \caption{MLLM with specialist decoder for multiple tokens.}
        \label{fig:introcom_sped}
    \end{subfigure}
    \\
    \begin{subfigure}{0.22\textwidth}
        \centering
        \includegraphics[width=0.75\linewidth]{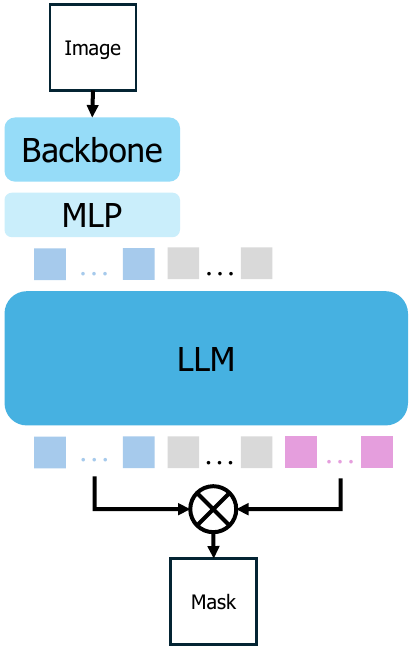}
        \caption{MLLM without a specialist decoder for multiple tokens.}
        \label{fig:introcom_ufo}
    \end{subfigure}
    \hfill 
    \begin{subfigure}{0.24\textwidth}
        \centering
        \includegraphics[width=0.75\linewidth]{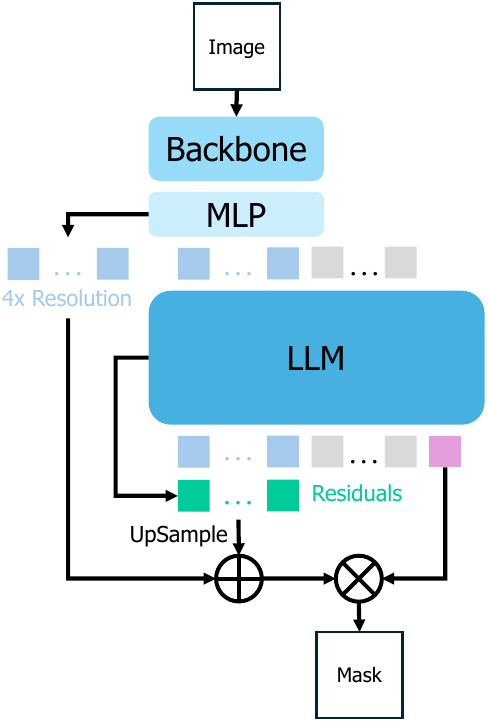}
        \caption{MLLM without a specialist decoder for a single token.}
        \label{fig:introcom_ours}
    \end{subfigure}
    \caption{Comparison of different MLLM-based segmentation paradigms. Almost all previous methods follow (a) and (b), which rely on the specialist mask decoder. 
    Limited approaches directly predict mask from the MLLM, yet they still require multiple \texttt{[SEG]} tokens for guidance. 
    Our approach in (d) takes advantage of higher resolution pre-compressed features and integrates the accumulated residual features, enabling MLLM-based segmentation without additional specialist decoders and \texttt{[SEG]} tokens.}
    \label{fig:introcom}
\end{figure}

With the rapid development of deep learning, especially the Large Language Models~(LLMs)~\cite{gpt320,llama23,qwen23,deepseekr125} and Multi-modal Large Language Models~(MLLMs)~\cite{llava23,qwen2vl24,internvl24}, the vision understanding evolves from classification of limited categories to more generalized and precise component explanations. 
Recently, several pioneering works~\cite{lisa24,glamm24,omgllava24} integrate specialist mask decoders~(\textit{e.g.}, Segment Anything Model~\cite{sam23}, Mask2Former~\cite{mask2former22}) into MLLMs, enabling challenging tasks like referring and reasoning segmentation.
These methods typically introduce a customized \texttt{[SEG]} token whose embedding prompts the mask decoder to generate precise masks from image features.
As shown in Fig.~\ref{fig:introcom_spe} and Fig.~\ref{fig:introcom_sped}, previous works typically rely on a specialized decoder that accepts segmentation-related tokens and visual features to generate high-quality masks, utilizing the powerful capabilities of pre-trained MLLMs and mask decoders.

Although previous studies have demonstrated the effectiveness of activating pre-trained mask decoders by a single \texttt{[SEG]} token from MLLMs, the introduction of additional parameters, as well as the complex structures, compromises the simplicity of the method and still results in a heavy reliance on external foundation models.
% these methods introduce additional parameters during both training and inference, which can reduce the model's efficiency and create a reliance on the performance of the mask decoder.
To address these limitations, recent work such as UFO~\cite{ufo25} explores decoder-free segmentation by replacing the mask decoder with a simple dot-product operation between image features and \texttt{[SEG]} tokens, both derived from the MLLM itself.
% generating segmentation masks solely within MLLMs.
% These approaches replace the redundant mask decoder with an attention-like dot product between image features and \texttt{[SEG]} tokens, both of which are derived from the MLLM. 
% Instead of relying on a specialized mask decoder, these approaches directly utilize the segmentation-related features from MLLM’s output. 
% Specifically, the segmentation features are projected into the same feature dimension as the visual representations extracted by MLLM. The similarity between these two modalities is then computed via dot-product attention, forming a small-scale similarity map. 
% Finally, this similarity map is interpolated to the original image resolution, where each pixel’s similarity score indicates the likelihood of belonging to the target mask. Through this approach, researchers have successfully developed mask-generator-free MLLM-based segmentation models, achieving promising results.
% anqi: Main downsampling is pixel shuffle, not patchify
% However, directly performing segmentation using MLLMs remains challenging. Since the visual encoder in MLLMs typically processes images by dividing them into patches and is constrained by a limited input sequence length, the resulting visual representations are significantly downsampled. 
However, this approach still encounters a fundamental challenge that most modern MLLMs~\cite{internvl325,qwen2vl24} incorporate pixel-shuffle downsampling with MLPs to spatially compressed visual features for efficient LLM processing, substantially reducing spatial resolution and losing fine-grained details critical for accurate segmentation.
% , in which the image features from the encoder are spatially compressed to facilitate more efficient understanding within the LLM. 
% This compression substantially reduces the feature resolution, inevitably leading to the loss of fine-grained spatial details that are critical for precise segmentation. 
% Consequently, such approaches often suffer from degraded mask quality and limited localization accuracy. 
UFO attempts to mitigate this by generating multiple \texttt{[SEG]} tokens to predict sub-pixel masks (Fig.~\ref{fig:introcom_ufo}), yet this compromise increases computational cost, whereas specialist decoders have shown that a single token is already sufficient.
% enhance segmentation resolution by predicting sub-pixels, yet the compromise solution increases computational overhead within LLM. 
% To mitigate these issues, UFO \cite{} utilizes multiple \texttt{[SEG]} tokens, each of which computes dot-product similarity with all image patches as shown in \ref{fig:introcom_ufo}. After calculating the similarity scores for each token, the corresponding segmentation masks are then aggregated and interpolated to the original image resolution. This approach helps to retain fine-grained spatial information and improves segmentation accuracy by combining the advantages of high-resolution feature mapping with the flexible reasoning capabilities of the MLLM.

This raises a critical question: \textit{Can we achieve high-quality segmentation using only a single \texttt{[SEG]} token without specialist decoders by better exploiting MLLM's intrinsic capabilities?} 
We answer with SELF1E (unlocking segmentation from MLLM it\textbf{SELF} with \textbf{1} \textbf{E}mbedding) by extracting high-resolution image features from MLLM's visual encoder to mitigate spatial loss caused by feature compression. 
Our key insight is that the resolution bottleneck stems not from single-token limitations but from information loss during pixel-shuffle compression. 
As illustrated in Fig.~\ref{fig:introcom_ours} and Fig.~\ref{fig:selfrep}, we preserve uncompressed image features from the image encoder's output features~(also referred to as pre-compressed features) by replicating each pixel according to the pixel-shuffle ratio.
% shows that our approach replicates each pixel of the output visual features by the image encoder~(also referred to as pre-compressed features) according to the pixel-shuffle ratio. 
This allows our model to obtain uncompressed features where spatial information is preserved, even after the pixel-shuffle process with MLP.
While compressed features proceed through the LLM as usual, we accumulate residual features from LLM layers, upsample them, and fuse them with the preserved uncompressed features.
% The compressed features for LLM still inherit the original setting, whereas the accumulated residual features from LLM are upsampled and fused with the uncompressed features. 
This Residual Features Refilling~(RFR) process effectively restores the resolution of fine-grained features with existing structures and features in MLLM. 
Moreover, to fully exploit high-resolution potential, we apply an MLP with pixel-unshuffle operations to both image features with and without LLM processing, respectively. 
% Moreover, to fully exploit high-resolution potential from visual features, we apply two MLPs with pixel-unshuffle modules for image features with and without LLM processing, respectively. 
These unshuffled features are employed in the Residual Features Amplifier~(RFA), enabling seamless fusion of LLM residual with uncompressed features at higher resolution. 
% Besides, the final integrated features are able to take the advantage of existing unshuffle module for even higher resolution. 
We further introduce a task-specific attention mask for the segmentation scenario, which facilitates the bidirectional interaction among all image features and \texttt{[SEG]} embeddings within LLM. 
% While our method primarily focuses on the high-resolution visual features extracted by the image encoder before inputting to the LLM, we do not discard the image features generated by the LLM itself. These features, extracted through the LLM, contain semantic information closely related to the textual instructions. This is particularly important for reasoning segmentation tasks, as it allows the visual features to contain more implicit, segmentation-related cues from the textual information.
% To leverage this, our approach incorporates the features generated by the LLM as residuals. Specifically, we compute the residual between the features before and after passing through the MLLMs, then project this residual features through a linear transformation. The processed residual features are added back to the high-resolution features extracted by the image encoder. This process allows us to simulate residual learning on high-resolution features with minimal additional computation. In doing so, we enhance the visual features with more segmentation-relevant information, enabling more precise segmentation at higher resolutions.
Extensive experiments demonstrate that our proposed model achieves substantial improvements in various tasks, including referring expression segmentation, reasoning segmentation, open-vocabulary segmentation, \etc. 

In summary, our contributions are as follows:
\begin{itemize}
    \item We propose an MLLM-based segmentation method that does not require a specialist mask decoder and multiple \texttt{[SEG]} tokens simultaneously for the first time. 
    % utilizes only a single segmentation token and does not rely on a mask decoder, yet still generates high-quality masks.
    % \item We leverage the image feature residuals provided by the LLM to optimize high-resolution visual features, thereby enhancing segmentation accuracy.
    \item We leverage original structures, uncompressed image features, and residual features from LLM to upgrade the resolution of fine-grained image features, which involves in RFR and RFA processes.   
    \item We specifically design the segmentation-specific attention mask to improve the bidirectional interaction between image features and \texttt{[SEG]} embeddings. 
    \item We demonstrate strong performance across multiple segmentation benchmarks while preserving VQA capabilities, validating that our approach maintains robust multi-granularity understanding.
    % Our model achieves promising results across multiple segmentation tasks and maintains high performance on VQA tasks, demonstrating robust image understanding capabilities across different granularities.
\end{itemize}

\section{Related Works}
\label{sec:rw}

% Early image segmentation models could only process a single image or rely on limited prompts (e.g., boxes or points) as input~\cite{sam23, mask2former22, maskrcnn17, deeplabv318, fcn15, unet15}, segmenting regions within a set of predefined categories.
% % While effective for fixed-class tasks, those models lack flexibility for diverse needs. 
% In contrast, multi-modal models that incorporate textual inputs allow more flexible and context-aware segmentation based on user instructions or specific scene descriptions - a capability much closer to real-world applications.

\subsection{Vision-Language Models for Segmentation}
The emergence of Vision-Language Models(VLMs) pretraining~\cite{align21, vilt21, alignfuse21, blip22, clip21, flava22} significantly narrows the gap between visual and textual modalities. This catalyzes the development of Open-Vocabulary Segmentation, which supersedes the traditional Semantic Segmentation methods~\cite{mask2former22, maskrcnn17, deeplabv318, fcn15, unet15,gfsam24,barm24,prformer25,combo25} that are restricted to predefined categories. 
% By leveraging their strong cross-modal understanding capacity, VLMs can be endowed with open-vocabulary segmentation and referring capabilities through efficient fine-tuning. ~\cite{openseg22, ovseg23, lseg22, clipseg22, denseclip22, groupvit22}
Methods like CLIPSeg~\cite{clipseg22} and LSeg~\cite{lseg22} utilize CLIP’s joint embedding space to make segmentation predictions from the similarity between pixel embeddings and textual embeddings. 
Some of the later methods~\cite{freeseg23,maskclip22}, \textit{e.g.}, ZegFormer~\cite{zegformer22}, OVSeg~\cite{ovseg23}, inherit the two-stage paradigm of MaskFormer series~\cite{maskformer21,mask2former22}, which first generate class-agnostic mask proposals, and then integrate the VLM to identify the category of each potential region. 
However, the repetitive encoding paradigm restricts the efficiency of such a paradigm. SAN~\cite{san23} designs a side-adapter to unify the two stages, which marks the diversion to one-stage methods~\cite{fcclip,sed24} with the VLM encoder. 
SCLIP~\cite{sclip24} further initiates the trend of modifying the structure of VLM~\cite{cliptrase24,cliper25} for segmentation without the need for finetuning. 
Moreover, the Referring Image Segmentation task further replaces the category description with natural language expressions. 
Considering the limitations of the CLIP text encoder in complex text understanding, most previous methods~\cite{gres23,cgformer23,magnet24}, \textit{e.g.} VLT~\cite{vlt21}, LAVT~\cite{lavt22}, incorporate attention between vision and language features derived from Swin-Transformer~\cite{swint21} and BERT~\cite{bert19}, respectively. 

% anqi: require more specific citations and descriptions
% Nevertheless, these approaches remain limited in complex language comprehension and reasoning ability, as their based models are primarily pre-trained on vision-language alignment tasks—such as matching images with their corresponding captions on datasets like Flickr30K~\cite{flickr15} and MSCOCO~\cite{mscoco15}—rather than on deeper reasoning or contextual understanding.
% This limitation stems from the fact that existing outdated models treat the language branch merely as a guidance signal instead of leveraging the strong comprehension and reasoning capabilities of MLLMs. To move beyond simple alignment, it is crucial to integrate the powerful reasoning capacity of MLLMs to deeper understand and interpret the relationships between visual features and textual instructions.

\subsection{MLLM-based Segmentation Models}
Recent studies have incorporated Multi-modal Large Language Models (MLLMs)~\cite{llava23, minigpt423, qwen224, flamingo24, internvl325} into segmentation tasks to enhance visual reasoning capabilities and achieve better alignment with complex visual-text semantics.
% Most existing methods couple an MLLM with a pre-trained mask decoder (e.g., SAM~\cite{sam23} or Mask2Former~\cite{mask2former22}) in a two-stage framework, where the MLLM first generates segmentation-related features that are subsequently processed by the mask decoder. 
Most existing methods introduce specialist segmentation foundation models~(e.g., SAM~\cite{sam23} or Mask2Former~\cite{mask2former22}) to decode the \texttt{[SEG]} token prompt from MLLMs. 
The pioneering work LISA~\cite{lisa24}, as well as several following approaches~\cite{cores25, glamm24, gsva24, lira25, omgllava24, sa2va25, visionllmv224, xsam25}, inputs both \texttt{[SEG]} tokens and visual features from an external encoder to the decoder. 
Some of the approaches~\cite{hyperseg24, pixelllm24, psalm25, rose25} get rid of the external encoder by applying multiple segmentation tokens and class-agnostic mask embeddings.
% Within this paradigm, there are two main approaches: one that inputs a single segmentation-related token (for each object) together with visual features into the decoder~\cite{lisa24, cores25, glamm24, gsva24, lira25, omgllava24, sa2va25, visionllmv224, xsam25}, and another that feeds multiple segmentation-related and mask-specific features into the decoder~\cite{hyperseg24, pixelllm24, psalm25, rose25}. 
% Although this approach generates high-quality segmentation results, it comes with the drawback of additional computational overhead due to the two-stage architecture. 
% HiMTok~\cite{himtok25} employ a lightweight mask de-tokenizer to alleviate complexity, but the model's architecture remains redundant and introduces additional parameters.
HiMTok~\cite{himtok25} employs a lightweight mask de-tokenizer that decodes the prediction mask from 32 tokens, thereby further removing image features from the decoder. 
Despite that, the specialist mask decoder still exists as an attachment. 
Recently, UFO~\cite{ufo25} became the first approach that simplifies the pipeline by discarding the mask decoder and predicting masks through dot-product similarity between \texttt{[SEG]} tokens and image embeddings that are both processed from MLLM, yet sacrifices the efficiency for generating 16 \texttt{[SEG]} tokens. 
% However, the need to generate multiple tokens still increases computational complexity and limits efficient utilization of high-resolution image features. 
% Our framework also discards the mask decoder and relies on a single segmentation token, while fully leveraging the high-resolution image features from the image encoder and the residual visual information from the MLLM.
Therefore, our paper discusses the feasibility of MLLM-based Segmentation without a specialist mask decoder and additional \texttt{[SEG]} tokens, and discovers how to activate more fine-grained features with better precision. 

\section{Methods}

\subsection{Overview}

We propose SELF1E, which performs MLLM-based segmentation without any specialist mask decoder. 
Our method follows the settings of a single additional segmentation token, yet directly produces the segmentation mask via matrix product~(Sec.~\ref{sec:pre}). 
In order to obtain high-resolution fine-grained image features, we first amplify the residual features from LLM~(Sec.~\ref{sec:nrps}), then combine them with the uncompressed image features with higher resolution from the image encoder~(Sec.~\ref{sec:nrpr}). 
Moreover, a further interaction among the segmentation token and image tokens is designed for better indication~(Sec.~\ref{sec:isti}). 

\subsection{Preliminaries}~\label{sec:pre}

% Almost all previous methods rely on a specialist mask generator. 
% The mask generator produces the prediction masks under the prompts of latent embeddings, which are generated from the MLLMs. 
% These methods, following the pioneer work LISA~\cite{lisa24}, define the label of this segmentation token as \texttt{[SEG]}. 
Following the pioneer work LISA~\cite{lisa24}, most of the methods define a \texttt{[SEG]} token to represent the latent embedding for segmentation. 
% Our method, although without a specialist mask generator, requires a \texttt{[SEG]} token for specifying the target as well. 
Our method inherit the Intern-VL~\cite{internvl325} series, which first obtain the original image features $\bm{F}_{V_0} \in \mathbb{R}^{\bm{N}_{\text{0}}\times d_0}$ of the image $x$ from the image encoder $\mathcal{E}$ belonging to the MLLM, then the image features $\bm{F}_{V_0}$ are adapted to compressed features $\bm{F}_{V_1}\in \mathbb{R}^{\bm{N}_{\text{1}}\times d}$ via pixel-shuffle with Multi Layer Perceptron~(MLP) by a factor of $\alpha > 1, \alpha \in \mathbb{N}^+$, \textit{i.e.}, $\bm{N}_{\text{0}} = \alpha \times \bm{N}_{\text{1}}$, while simultaneously projecting the token dimension from $\alpha d_0$ to $d$. 
The combination of $\bm{F}_{V_1}$ and other segmentation-guidance text embeddings is fed into the LLM $\mathcal{M}$, producing the \texttt{[SEG]} token $\bm{F}_{\text{SEG}}$. Meanwhile, the latent tokens corresponding to the image positions are gathered as the LLM-processed \texttt{[IMG]} tokens $\bm{F}_{\text{IMG}}$. 
Refer to the previous methods~\cite{ufo25} without a specialist mask decoder, we produce the segmentation mask $\hat{y}\in \mathbb{R}^{\bm{N}_{\text{IMG}} \times \bm{N}_{\text{SEG}}}$ from $\bm{N}_{\text{IMG}}$ post-processed image tokens $\bm{F}_{\text{IMG}}' $and $\bm{N}_{\text{SEG}}$ post-processed \texttt{[SEG]} tokens $\bm{F}_{\text{SEG}}'$ by:
\begin{equation}\label{eq:seg}
    \bm{\hat{y}} = \frac{\bm{F}_{\text{IMG}}' \bm{F}_{\text{SEG}}'^\top}{\sqrt{d}},
\end{equation}
where $d$ represents the dimension of latent embeddings. 

\subsection{Residual Features Refilling}\label{sec:nrpr}

\begin{figure}
    \centering
    \includegraphics[width=0.9\linewidth]{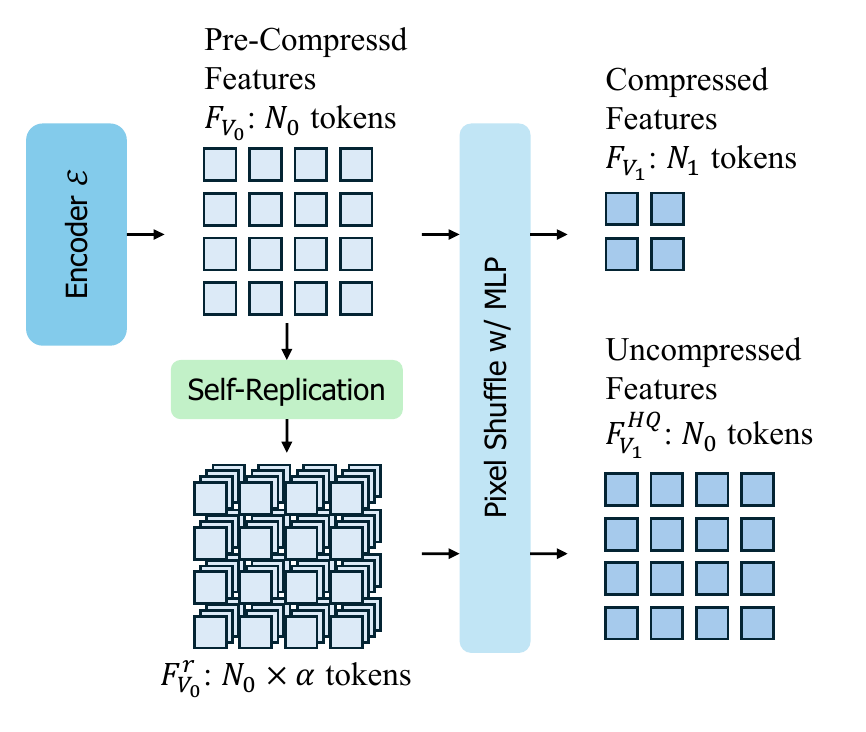}
    \caption{The additional branch of pre-compressed image features self-replication for uncompressed features. The compressed features for LLM follow the original process. }
    \label{fig:selfrep}
\end{figure}

% Within most modern MLLMs, a pixel-shuffle MLP $f_{MLP}$ is employed to downsample the spatial resolution of the feature map from $\bm{F}_{V_0}$ to $\bm{F}_{V_1}$ by a factor of $\alpha$, \textit{i.e.}, $\bm{N}_{\text{0}} = \alpha \times \bm{N}_{\text{1}}$, while simultaneously projecting the token dimension to $d$. 
% This process improves the overall efficiency, yet reduces the resolution of the image features.  
The image features compression in modern MLLMs improves the efficiency of visual understanding, whereas the reduced resolution becomes a bottleneck for segmentation tasks. 
Prior works without specialist mask decoders directly predict masks from $\bm{F}_{\text{IMG}}\in \mathbb{R}^{\bm{N}_{\text{1}}\times d}$ that under the compressed resolution.~\cite{ufo25} 
In contrast, high-resolution image representations primarily enlarge the potential for more precise mask prediction, \textit{e.g.}, methods with SAM~\cite{sam23} exploit features with much higher resolution compared to $\bm{F}_{\text{IMG}}$. 

Therefore, we focus on image features $\bm{F}_{V_0}$, which possess a relatively higher resolution of $\bm{N}_0$.
Different from the inherent pixel-shuffle with MLP operation that compresses features to a lower resolution of $\bm{N}_1$, we extend an additional branch for uncompressed features presented in Fig.~\ref{fig:selfrep}. 
Specifically, each pixel feature is replicated $\alpha$ times to form $\bm{F}_{V_0}^r \in \mathbb{R}^{N_0\times\alpha d}$, and then apply the same compression MLP to each expanded pixel. 
This procedure conducts the intra-pixel compression and produces $\bm{F}_{V_1}^{HQ}\in \mathbb{R}^{\bm{N}_{\text{0}}\times d}$, thereby preserving the original spatial resolution. 
% The common pixel-shuffle process could compress the spatial resolution as $\bm{F}_{V_1}$ for LLM. 
% However, to maintain the spatial resolution, we first replicate the feature of each pixel $\alpha$ times as $\bm{F}_{V_0}^r \in \mathbb{R}^{N_0\times\alpha d}$, then apply the same MLP $f_{MLP}$ to simulate the pixel-shuffle process and generate $\bm{F}_{V_1}^{HQ}\in \mathbb{R}^{\bm{N}_{\text{0}}\times d}$. 
Considering that the features of a specific pixel are largely similar to those of its neighboring pre-shuffled pixels, the replication of features enables an approximate simulation of the pre-shuffled features corresponding to each pixel. 

Compared to the $\bm{F}_{\text{IMG}}$ with compressed resolution, the uncompressed image features $\bm{F}_{V_1}^{HQ}$ offer a significant advantage in generating more precise segmentation masks following Eq.~\ref{eq:seg}. 
However, the uncompressed features from the encoder are insufficient for fine-grained conceptual segmentation, as their representations mainly capture the category-level semantics. 
These features often exhibit limited distinctiveness compared with $\bm{F}_{\text{IMG}}$ that is refined by the LLM. 
% The $\bm{F}_{V_1}^{HQ}$ lacks further identification of fine-grained features, while $\bm{F}_{\text{IMG}}$ from LLM lacks pixels for better resolution. 
The optimal design should leverage the resolution advantage of $\bm{F}_{V_1}^{HQ}$ while integrating the semantic granularity advantage of $\bm{F}_{\text{IMG}}$. 
The compressed features $\bm{F}_{V_1}$ are able to serve as the connector, where $\bm{F}_{V_1}^{HQ}$ represents an expanded version of $\bm{F}_{V_1}$, and $\bm{F}_{\text{IMG}}$ is derived from $\bm{F}_{V_1}$ through LLM processing. 
% Thus, we accumulate all the residual features $\bm{F}_l^\text{NR}$ from every transformer block at the layer $l$ as $\bm{F}^\text{NR}$ via:
Thus, we accumulate the overall residuals $\bm{F}_\text{R}\in \mathbb{R}^{\bm{N}_{\text{1}}\times d}$ from LLM via:
\begin{equation}\label{eq:nr}
    \bm{F}_\text{R} = \bm{F}_{\text{IMG}} - \bm{F}_{V_1},
\end{equation}
then simply upsample the residual features and add them to the $\bm{F}_{V_1}^{HQ}$:
\begin{equation}
    \bm{F}_{\text{IMG}}' = \bm{F}_{V_1}^{HQ} + \mathcal{I}(\bm{F}_\text{R}), 
\end{equation}
where $\mathcal{I}$ refers to the upsampling operation in the spatial dimension with a factor of $\alpha$. 
Following Eq.~\ref{eq:seg}, fused features $ \bm{F}_{\text{IMG}}'\in \mathbb{R}^{\bm{N}_{\text{0}}\times d}$ can generate a prediction mask with higher resolution, while integrating fine-grained distinctiveness. 

\begin{figure}[!t]
    \centering
    \includegraphics[width=\linewidth]{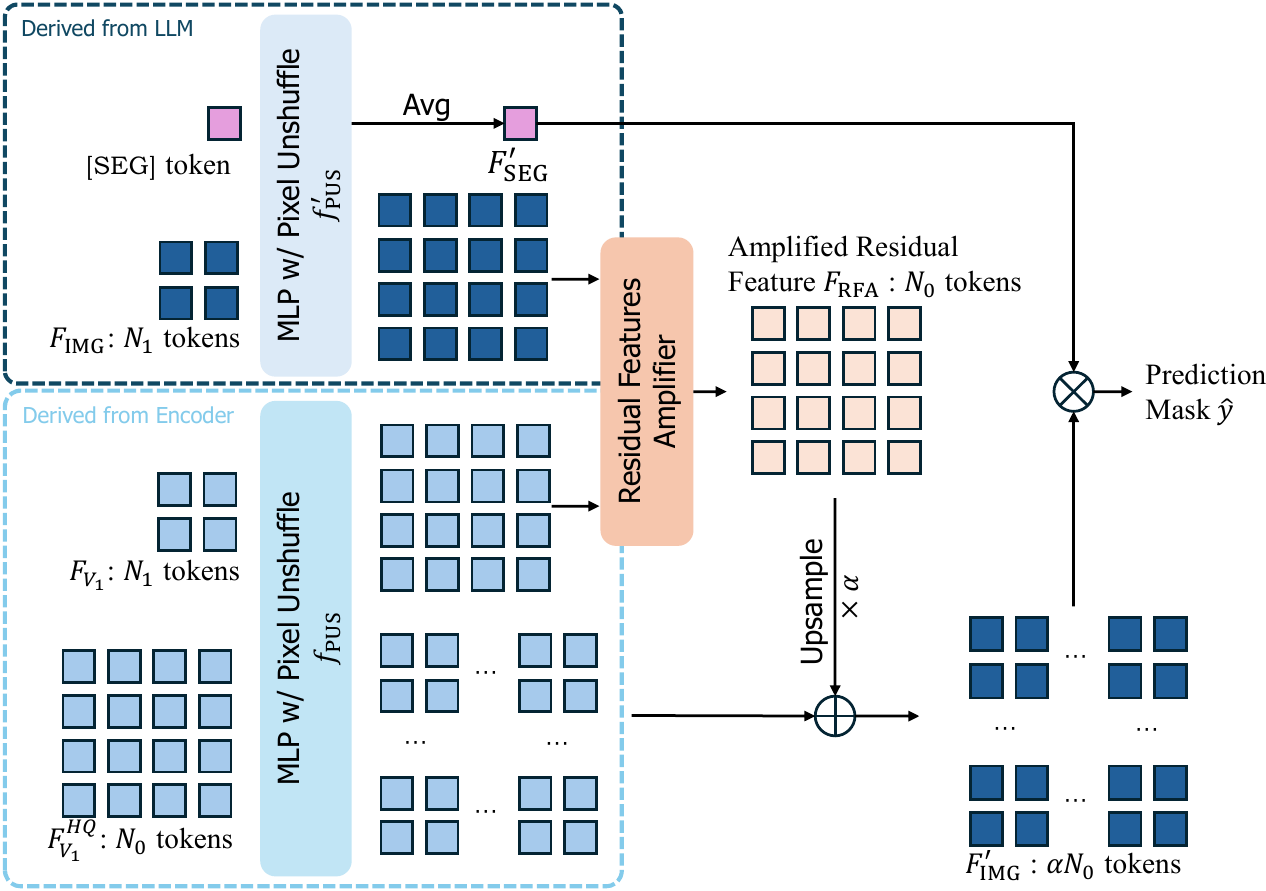}
    \caption{An overview of RFR and RFA operations. The residual features are amplified from the restored compressed features with and without LLM processing. The fusion of restored uncompressed features and the amplified residual features simultaneously achieves higher resolution and fine-grained representations.}
    \label{fig:nrstruct}
\end{figure}

\subsection{Residual Features Amplifier}\label{sec:nrps}

The RFR operation allows the coexistence of higher resolution and finer semantic granularity within the image features from MLLM. 
However, the features once processed by the Pixel-Shuffle-with-MLP still implicitly contain information from higher resolution, as each embedding corresponds to features from $\alpha$ pixels. 
We introduce an MLP-with-Pixel-Unshuffle operation $f_{\text{PUS}}$ to restore the compressed feature representations. 

Despite the resolution improvement brought by the Pixel-Unshuffle process, the direct combination of $\bm{F}_{V_1}^{HQ}$ and $\mathcal{I}(\bm{F}_\text{R})$ does not achieve seamless semantic alignment with the compressed features. 
The uncompressed features $\bm{F}_{V_1}^{HQ}$ contains $\bm{N}_{0}$ pixels, whereas $\mathcal{I}(\bm{F}_\text{R})$ is originally interpolated from $\bm{F}_\text{R}$, which has $\alpha^2$ times fewer pixels than the unshuffled features. 
Besides, the Pixel-Shuffle with MLP operation is applied to the formal image features, while the residual features $\bm{F}_\text{R}$ are intermediate representations that are insufficient for direct processing through MLP with Pixel-Unshuffle. 
Therefore, as shown in Fig.~\ref{fig:nrstruct}, we revisit Eq.~\ref{eq:nr} and introduce the Residual Features Amplifier operation as follows:
\begin{equation}
    \bm{F}_\text{RFA} = f_{\text{PUS}}'(\bm{F}_{\text{IMG}}) - f_{\text{PUS}}(\bm{F}_{V_1}),
\end{equation}
where both $\bm{F}_{V_1}$ and $\bm{F}_{\text{IMG}}$ fit the requirement of compressed image features. 
The two MLPs with Pixel-Unshuffle functions, $f_{\text{PUS}}$ and $f_{\text{PUS}}'$, are applied to the image features without and with process from LLM, respectively. 
The amplified Residual Features $\bm{F}_\text{RFA} \in \mathbb{R}^{N_0\times d}$ has the same dimensionality as $\bm{F}_{V_1}^{HQ}$ and $\bm{F}_{V_0}$. 
% Considering that $f_{\text{PUS}}'$ and $f_{\text{PUS}}$ is utilized for image features w/ and w/o LLM process, respectively, we are able to take advantage of $f_{\text{PUS}}'$ for further upsampling:
Since the derivation from the $\bm{F}_{V_0}$ to $\bm{F}_{V_1}^{HQ}$ involves self-replication, and considering the requirements of feature alignment through $f_{\text{PUS}}$, the combination of $\bm{F}_{V_1}^{HQ}$ and $\bm{F}_\text{RFA}$ is more seamless from the original resolution of $\bm{N}_0$. 
Thus, we fuse the $\bm{F}_{V_1}^{HQ}$ and $\bm{F}_\text{RFA}$ as $\bm{F}_{\text{IMG}}' \in \mathbb{R}^{\alpha \bm{N}_{\text{0}}\times d}$:
\begin{equation}
    \bm{F}_{\text{IMG}}' = f_{\text{PUS}}(\bm{F}_{V_1}^{HQ}) + \mathcal{I}(\bm{F}_\text{RFA}). 
\end{equation}
% where the image features finally reach $\alpha^2$ times resolution of original $F_{IMG}$ from LLM with assistance from existing higher resolution features and structures. 
Furthermore, the \texttt{[SEG]} embedding $\bm{F}_{\text{SEG}}$ is post-processed with $f_{PUS}'$ for further alignment:
\begin{equation}
    % \bm{F}_{\text{SEG}}' = \operatorname{mean} f_{\text{PUS}}'(\bm{F}_{\text{SEG}})], 
    \bm{F}_{\text{SEG}}' = \operatorname{mean}\!\big(f_{\text{PUS}}'(\bm{F}_{\text{SEG}})\big),
\end{equation}
where $\operatorname{mean}(\cdot)$ denotes the averaging function over the $\alpha$ unshuffled embeddings corresponding to the \texttt{[SEG]} token. 

\begin{table*}[!t]
    \centering
    \resizebox{0.85\linewidth}{!}{
        \begin{tabular}{r|cc|ccc|ccc|cc}
        \toprule[1.1pt] 
        \multirow{2}{*}{ Method } & \multirow{2}{*}{ w/o SMD } & \multirow{2}{*}{1-Token} & \multicolumn{3}{c|}{ RefCOCO } & \multicolumn{3}{c|}{ RefCOCO+ } & \multicolumn{2}{c}{ RefCOCOg }\\
        % \cline { 4 - 11 }
        & & & val & testA & testB & val & testA & testB & val & test \\
        \midrule[0.7pt]
        LISA-7B(ft)~\cite{lisa24} & $\times$ & $\checkmark$ & 74.9 & 79.1 & 72.3 & 65.1 & 70.8 & 58.1 &  67.9 & 70.6 \\
        PixelLM-7B~\cite{pixelllm24} & $\times$ & $\checkmark$ & 73.0 &76.5 & 68.2 & 66.3 & 71.7 & 58.3 &69.3 & 70.5 \\
        GSVA-7B~\cite{gsva24} & $\times$ & $\checkmark$ & 76.4 &77.4 &72.8 &64.5 &67.7 & 58.6 & 71.1 & 72.0 \\
        GSVA-7B(ft)~\cite{gsva24} & $\times$ & $\checkmark$ & 77.2 &78.9 &73.5 &65.9 &69.6 & 59.8 & 72.7 & 73.3 \\
        u-LLaVA~\cite{ullava23} & $\times$  & $\checkmark$ & 83.0   & 85.1    & 80.5  & 77.1  & 81.7  & 70.6    & 77.1  & 78.0 \\
        LaSagnA-7B~\cite{lasagna24} & $\times$ & $\checkmark$ & 76.8 & 78.7 & 73.8 & 66.4 & 70.6 & 60.1 & 70.6 & 71.9 \\
        OMG-LLaVA ~\cite{omgllava24} & $\times$ & $\checkmark$ & 75.6   & 77.7 & 71.2   & 65.6    & 69.7     & 58.9   & 70.7     & 70.2  \\
        OMG-LLaVA(ft) ~\cite{omgllava24} & $\times$ & $\checkmark$ & 78.0   & 80.3  & 74.1   & 69.1    & 73.1     & 63.0   & 72.9     & 72.9  \\
        GLaMM~\cite{glamm24} & $\times$ & $\checkmark$ & 79.5    & 83.2    & 76.9  & 72.6    & 78.7     & 64.6    &  74.2   & 74.9  \\
        VisionLLM v2~\cite{visionllmv224} & $\times$ & $\checkmark$ & 76.6 & 79.3 & 74.3 & 64.5 & 69.8 & 61.5 & 70.7 & 71.2 \\
        PSALM ~\cite{psalm25} & $\times$ & $\times$ & 83.6   & 84.7  & 81.6   & 72.9    & 75.5     & 70.1   & 73.8     & 74.4 \\ 
        GroundHog-7B~\cite{groundhog24} & $\times$ & $\times$ & 78.5 & 79.9 & 75.7 & 70.5 & 75.0 & 64.9 & 74.1 & 74.6 \\
        SAM4MLLM-8B~\cite{sam4mllm24} & $\times$ & - & 79.8 & 82.7 & 74.7 & 74.6 & 80.0 & 67.2 & 75.5 & 76.4 \\
        HyperSeg~\cite{hyperseg24} & $\times$ & $\times$ & \underline{84.8} & \underline{85.7} & \underline{83.4} & 79.0 & \underline{83.5} & 75.2 & 79.4 & 78.9 \\
        % \hline
        HiMTok-8B~\cite{himtok25} &$\times$ &  $\times$ & 81.1 & 81.2 & 79.2 & 77.1 & 78.8 & 71.5 & 75.8 & 76.7 \\
        HiMTok-8B(ft)~\cite{himtok25} &$\times$ &  $\times$ & \textbf{85.0} & 85.2 & \textbf{83.5} & \underline{79.7} & 82.7 & \underline{76.0} & 80.0 & 80.6 \\
        UFO-8B~\cite{ufo25} & $\checkmark$ &  $\times$ & 80.0 & 81.6 & 78.1 & 76.7 & 79.9 & 72.3 & 75.5 & 76.3 \\
        UFO-8B(ft)~\cite{ufo25} & $\checkmark$ &  $\times$ & 81.0 & 82.6 & 78.6 & 77.1 & 80.4 & 72.6 & 76.7 & 77.3 \\
        \midrule
        SELF1E-2B & $\checkmark$ &  $\checkmark$ & 80.2 & 82.1 & 77.6 & 74.6 & 79.1 & 69.2 & 77.0 & 77.8 \\
        SELF1E-SEG-2B &  $\checkmark$ &  $\checkmark$ & 84.3 & 85.4 & 82.3 & 78.9 & \underline{83.5} & 75.1 & \underline{80.4} & \underline{81.0} \\
        SELF1E-8B & $\checkmark$ &  $\checkmark$ & 82.5 & 83.8 & 80.1 & 77.6 & 81.6 & 73.7 & 79.1 & 79.8 \\
        SELF1E-SEG-8B & $\checkmark$ &  $\checkmark$ & \underline{84.7} & \textbf{86.2} & \underline{83.4} & \textbf{80.2} & \textbf{84.2} & \textbf{77.0} & \textbf{82.1} & \textbf{82.8} \\
        \bottomrule
        \end{tabular}
    }
    \caption{Comparison of cIoU on the Referring Expression Segmentation benchmarks (RefCOCO/+/g). 
    SMD represents Specialist Mask Decoder, and 1-Token represents using a single special token for segmentation. 
    ``ft'' denotes the model is finetuned on the specific dataset.
    Results in \textbf{bold} are the best, while \underline{underlined} are the second best. }
    \label{tab:res_res}
\end{table*}

\subsection{Image-Segmentation Token Interaction}~\label{sec:isti}

Most previous methods, whether they follow the causal inference paradigm or not, rarely discuss the interaction between the \texttt{[IMG]} tokens and \texttt{[SEG]} token. 
However, the settings of \texttt{[CLS]} tokens in ViT~\cite{vit20} and prompt tokens in SAM~\cite{sam23} demonstrate the effectiveness of bidirectional interaction between different kinds of embeddings, where these tokens with special purpose could gain and upgrade the knowledge of image features during each interaction. 

Inspired by these approaches, we specifically design an attention mask for segmentation purposes. 
We set a bidirectional attention mask among the positions of \texttt{[IMG]} tokens, so that the understanding of image features is not restricted to the sequential position. 
Then, the \texttt{[SEG]} tokens are set to be visible to the \texttt{[IMG]} tokens, enabling a bidirectional perception whenever the \texttt{[SEG]} token is presented for segmentation purposes. 
The redesigned attention mask for segmentation facilitates the bidirectional interaction between \texttt{[SEG]} token and \texttt{[IMG]} tokens, while the original causal inference ability is still maintained for VQA purposes. 

\subsection{Training Objectives}

The training process of our approach involves both VQA samples and segmentation samples with specific templates. 
The cross-entropy loss $\mathcal{L}_{\text{text}}$ for autoregressive text prediction follows the original setting of MLLMs. 
The prediction masks for segmentation purpose require pixel-level cross-entropy loss $\mathcal{L}_{\text{BCE}}$ and DICE loss $\mathcal{L}_{\text{DICE}}$ for optimization. Overall, the summarized loss $\mathcal{L} = \mathcal{L}_{\text{text}} + \mathcal{L}_{\text{BCE}} +\mathcal{L}_{\text{DICE}}$ . 
\section{Experiments}

\subsection{Implementation Details}

Our approach uses InternVL3-2B/8B~\cite{internvl325} as the base model without any additional specialist mask decoders. 
The training process involves multiple datasets following the previous methods~\cite{lisa24}, including ADE20k~\cite{ade20k18}, COCOStuff~\cite{cocostuff18}, Pascal-Part~\cite{pascal_part14}, and LVIS-PACO~\cite{paco23} for semantic segmentation, RefCOCO, RefCOCO+ and RefCOCOg~\cite{refcoco/+14, refcoco/+/g16} for refering segmentation, ReasonSeg~\cite{lisa24} for reasoning segmentation, and several VQA datasets~\cite{llava23,textvqa19,vqav217,vizwiz18,okvqa19,gqa19} for maintaining the original VQA ability.  
The finetuning of the MLLM requires LoRA~\cite{lora22} adaptation with a rank of 128 for SELF1E-2B and 64 for SELF1E-8B. 
The learning rate of finetuning is set to 1e-4 for SELF1E-2B and 6e-5 for SELF1E-8B with the AdamW optimizer and a cosine scheduler. 
The whole training process is conducted on NVIDIA A100 / RTX4090 GPUs with an overall gradient accumulated batch size of 160. 
As most prior works adopt different training strategies to emphasize specific capabilities, we employ two strategies for a comprehensive comparison. 
The vanilla SELF1E version requires 1 epoch of training with all selected samples from the datasets, while the SELF1E-SEG version employs different sampling frequencies for each data type to emphasize segmentation performance.
The details of the datasets, training settings, and VQA experimental results are shown in the Appendix. 

\subsection{Comparison with State-of-the-arts}

\subsubsection{Referring Expression Segmentation}
Referring Expression Segmentation (RES) is a canonical benchmark for evaluating language-guided segmentation, where the models are required to localize and segment the target object described by a natural language expression. 
We conduct experiments on three standard RES benchmarks: RefCOCO, RefCOCO+, and RefCOCOg~\cite{refcoco/+14, refcoco/+/g16} following the evaluation protocol using the cIoU metric.

As shown in Tab.~\ref{tab:res_res}, our proposed SELF1E-SEG-2B surpasses most of the prior SOTA across all benchmarks, achieving 85.4\% cIoU on RefCOCO testA, 83.5\% on RefCOCO+ testA, and 80.4\% and 81.0\% on RefCOCOg val and test, respectively. The large-scale version SELF1E-SEG-8B further pushes performance boundaries, with improvements of 0.5\% on RefCOCO testA, 0.7\% on RefCOCO+ testA, and 2.2\% on RefCOCOg test, outperforming recent high-performing approaches such as HiMTok-8B(ft)~\cite{himtok25} and UFO-8B(ft)~\cite{ufo25}.
% Our design simplifies the architecture of MLLM-based segmentation models, thereby reducing interference among multiple segmentation-related tokens and eliminating the need for additional alignment between the MLLM and the mask generator. This streamlined 
These results validate the effectiveness of our SELF1E without a specialist mask decoder, which showcases the segmentation ability solely from MLLM. 
% Our design effectively enhances the performance on RES tasks. 
Notably, even without emphasizing fine-tuning on segmentation samples, our model achieves competitive results, highlighting its inherent segmentation potential from the original MLLM. 
% highlighting its balance between accuracy and efficiency across diverse tasks.

\newcommand{\graycell}[1]{\textcolor{gray}{#1}}
\begin{table}[t]
    \centering
    \resizebox{\linewidth}{!}{
        \begin{tabular}{r|c|cc|cc|cc}
        \toprule 
        \multirow{2}{*}{Method} & \multirow{2}{*}{ZS} & \multicolumn{2}{c|}{ val } & \multicolumn{2}{c|}{ testA } & \multicolumn{2}{c}{ testB } \\
        % \cline { 3 - 8} 
        & & cIoU & gIoU & cIoU & gIoU & cIoU & gIoU \\
        \midrule
        LISA-7B\cite{lisa24}  & $\times$ & \graycell{38.7} & \graycell{32.2} & \graycell{52.6} & \graycell{48.5} & \graycell{44.8} &  \graycell{39.7} \\
        LISA-7B(ft)\cite{lisa24}  & $\times$ & \graycell{61.8} & \graycell{61.6} & \graycell{68.5} & \graycell{66.3} & \graycell{60.6} & \graycell{58.8} \\
        GSVA-7B\cite{gsva24} & $\times$ & \graycell{61.7} &\graycell{63.3} &\graycell{69.2} &\graycell{70.1} & \graycell{60.3} & \graycell{61.3} \\
        GSVA-7B(ft)\cite{gsva24} & $\times$ & \graycell{63.3} &\graycell{66.5} &\graycell{69.9} &\graycell{71.1} & \graycell{60.5} & \graycell{62.2} \\
        SAM4MLLM-8B~\cite{sam4mllm24} & $\times$ & \graycell{67.8} &\graycell{71.9} &\graycell{72.2} &\graycell{74.2} & \graycell{63.4} & \graycell{65.3} \\
        HiMTok-8B~\cite{himtok25} & $\times$ & \graycell{66.8} & \graycell{68.7} & \graycell{68.6} & \graycell{67.6} & \graycell{65.8} & \graycell{64.1} \\
        \midrule
        LaSagnA~\cite{lasagna24} & \checkmark  & 38.1 & 32.4 & 50.4 & 47.3 & 42.1 & 38.9 \\
        PSALM~\cite{psalm25} & \checkmark & \underline{42.0} &\textbf{43.3} &52.4 &\textbf{54.5} & \underline{50.6} & \textbf{52.5} \\
        OMG-LLaVA~\cite{omgllava24} & \checkmark & 39.3 & 36.1 & 52.4 & 50.1 & 43.7 & 42.2 \\
        LIRA-8B~\cite{lira25} & \checkmark & 40.9 & 36.7 & 52.4 & 50.4 & 44.9 & 42.4  \\
        \midrule  
        SELF1E-2B & \checkmark & 41.4 & 35.2 & \underline{55.6} & 51.4 & 47.0 & 42.7 \\
        SELF1E-8B & \checkmark & \textbf{44.4} & \underline{37.5} & \textbf{57.5} & \underline{53.2} & \textbf{50.9} & \underline{45.6} \\
        \bottomrule
        \end{tabular}
    }
    \caption{Comparison on Generalized Referring Expression Segmentation. ZS denotes whether the method is zero-shot.}
    \label{tab:grefcoco}
\end{table}

\subsubsection{Generalized Referring Expression Segmentation}

We evaluate our approach on the gRefCOCO~\cite{grefcoco23} dataset, which is designed for Generalized Referring Expression Segmentation~(GRES) that contains scenarios of referring to multiple target objects as well as nonexistent objects. 
We follow the evaluation template of RES and conduct the zero-shot inference without any dataset-specific fine-tuning. 
As illustrated in Tab.~\ref{tab:grefcoco}, our approach achieves superior performance compared to all existing zero-shot methods. 
In particular, SELF1E-8B achieves 44.4\% / 37.5\% (cIoU / gIoU) on the validation split, 57.5\% / 53.2\% on testA, and 50.9\% / 45.6\% on testB, demonstrating state-of-the-art performance in cIoU across each subset. 

\begin{table}[t]
    \centering
    \resizebox{0.9\linewidth}{!}{
        \begin{tabular}{r|cc|cc}
        \toprule 
        \multirow{2}{*}{ Method } & \multicolumn{2}{c|}{ val } & \multicolumn{2}{c}{ test } \\
        \cline { 2 - 5 } & gIoU & cIoU & gIoU & cIoU \\
        \midrule
        LISA-7B\cite{lisa24}  & 44.4 & 46.0 & 36.8 & 34.1 \\
        LISA-7B(ft)\cite{lisa24}  & 52.9 & 54.0 & 47.3 & 48.4 \\
        % GroundHog-7B~\cite{groundhog24} & 56.2 &- &- &- \\
        % VisionLLM v2~\cite{visionllmv224}  & 51.0 & - & - & - \\
        LaSagnA~\cite{lasagna24}  & 48.8 & 47.2 & - & - \\
        VISA-7B~\cite{visa24} & 52.7 & 57.8 & - & - \\
        SAM4MLLM-8B~\cite{sam4mllm24} & 58.4 &60.4 &- &- \\
        CoReS-7B~\cite{cores25} & 59.4 & - & 52.4 & - \\
        \midrule
        HiMTok-8B & 60.7 & 67.0 & 60.8 & 66.2 \\
        UFO-8B & 60.0 & - & - & - \\
        SELF1E-2B & 59.4 & 68.8 & 57.3 & 57.6 \\
        SELF1E-8B & \textbf{65.9} & \textbf{69.7} & \textbf{65.7}& \textbf{67.0} \\
        \bottomrule
        \end{tabular}
    }
    \caption{Comparison of our approach and other state-of-the-art methods on Reasoning Segmentation.}
    \label{tab:reasonseg}
\end{table}

\subsubsection{Reasoning Segmentation}
Reasoning Segmentation, introduced by LISA~\cite{lisa24}, is a challenging benchmark for reasoning-driven segmentation, where models must interpret complex and indirect linguistic instructions and perform multi-step reasoning grounded in world knowledge. 
% We evaluate our model based on the InternVL3 backbone at both 2B and 8B scales. 
Remarkably, as shown in Table~\ref{tab:reasonseg}, even our SELF1E-2B model achieves performance comparable to strong baselines such as HiMTok and UFO. 
It surpasses the previous SOTA by 1.8\% in cIoU, demonstrating that our approach is able to extract higher resolution fine-grained features for segmentation even with a lightweight model capacity. 
The large-scale SELF1E-8B further establishes new SOTA results across all metrics, achieving gIoU advantage of 5.2\%~(val) and 4.9\%~(test), and cIoU advantage of 2.7\%~(val) and 0.8\%~(test). 
The results show that even after fine-tuning with visual tokens, the MLLM retains its ability to understand and reason over complex linguistic instructions.

% We attribute these improvements to our NRPR and NRPS modules. NRPR extracts residual image features from the LLM that align with reasoning instructions, capturing fine-grained information, while NRPS simulates the high-resolution features before the Pixel-Shuffle process to enhance spatial detail. Combining these with the visual encoder features ensures that the resulting image features retain both high precision and high resolution, which is crucial for reasoning-driven segmentation. 
% Furthermore, using a single segmentation token prevents feature entanglement that can arise when multiple segmentation-related tokens are used. Overall, these contributions enable our model to possess strong reasoning capabilities while effectively combining them with high-quality segmentation performance (see ablation studies in Sec.~\ref{sec:} for detailed analysis of NRPR and NRPS).

% We attribute these improvements to the RFR and RFA modules. The RFR module captures fine-grained residual image features from the LLM that are aligned with reasoning instructions, while the RFA module enhances spatial details by generating high-resolution features. Together, these designs strengthen the model’s reasoning ability and segmentation quality (see ablation results in Sec.~\ref{sec:abl_rfr} and Sec.~\ref{sec:abl_rfs}).

\subsubsection{Open-Vocabulary Segmentation}
We also evaluate our model on the open-vocabulary segmentation (OVS) task, which involves segmenting previously unseen categories. 
Generating all masks simultaneously is impractical for datasets with a large number of categories, and requiring the model to produce numerous \texttt{[SEG]} tokens per prompt may lead to semantic ambiguity. 
To address this, we query each category individually, producing one mask per query, and assign each pixel to the category with the highest similarity score.

\begin{table}[!t]
    \centering
    \begin{tabular}{r|cccc}
    \toprule
          & A150 & PC59 & PC459 & PAS20 \\
    \midrule
    PSALM~\cite{psalm25} & 18.2 & 48.5 & 10.2 & 81.3 \\    
    HiMTok-8B~\cite{himtok25} & 25.0 & 43.9 & - & 82.0 \\
    HyperSeg~\cite{hyperseg24} & 22.3 & 64.6 & - & 92.1 \\
    % ROSE-7B~\cite{rose25} & 51.0 & - & - & - \\
    \midrule
    SELF1E-2B & 50.2 & 64.6 & 42.4 & 89.9 \\
    SELF1E-2B* & 24.0 & 36.9 & 18.9 & 44.8 \\
    \bottomrule
    \end{tabular}
    \caption{Comparison with SOTA methods on open-vocabulary semantic segmentation benchmarks. 
    We use mIoU as the evaluation metric for semantic segmentation. 
    The datasets are abbreviated as: ADE20k-150 (A150), Pascal Context-59 (PC59), Pascal Context-459 (PC459), Pascal VOC-20 (PAS20). 
    ``*'' denotes that the model is trained without any data from the corresponding datasets. }
    \label{tab:ovs}
\end{table}

Results on ADE20k-150~\cite{ade20k18}, Pascal Context-59~\cite{pascal_context14}, Pascal Context-459~\cite{pascal_context14}, and Pascal VOC-20~\cite{pascal_voc10} are shown in Tab.~\ref{tab:ovs}, with mean Intersection-over-Union (mIoU) as the evaluation metric. 
Our SELF1E achieves strong performance across all benchmarks, reaching SOTA levels on ADE20k and notable gains 42.4\% on Pascal Context-459. The SELF1E* version was trained fully on RefCOCO-series datasets. 
When evaluated on OVS tasks in a zero-shot manner, it still performs competitively, demonstrating the model’s robust generalization ability.
Considering that the related training datasets used by other models are uncertain, our results demonstrate the effectiveness of the proposed model on OVS tasks, particularly on large-scale datasets with diverse category distributions.

\begin{figure*}[!t]
    \centering
    \includegraphics[width=0.87\textwidth]{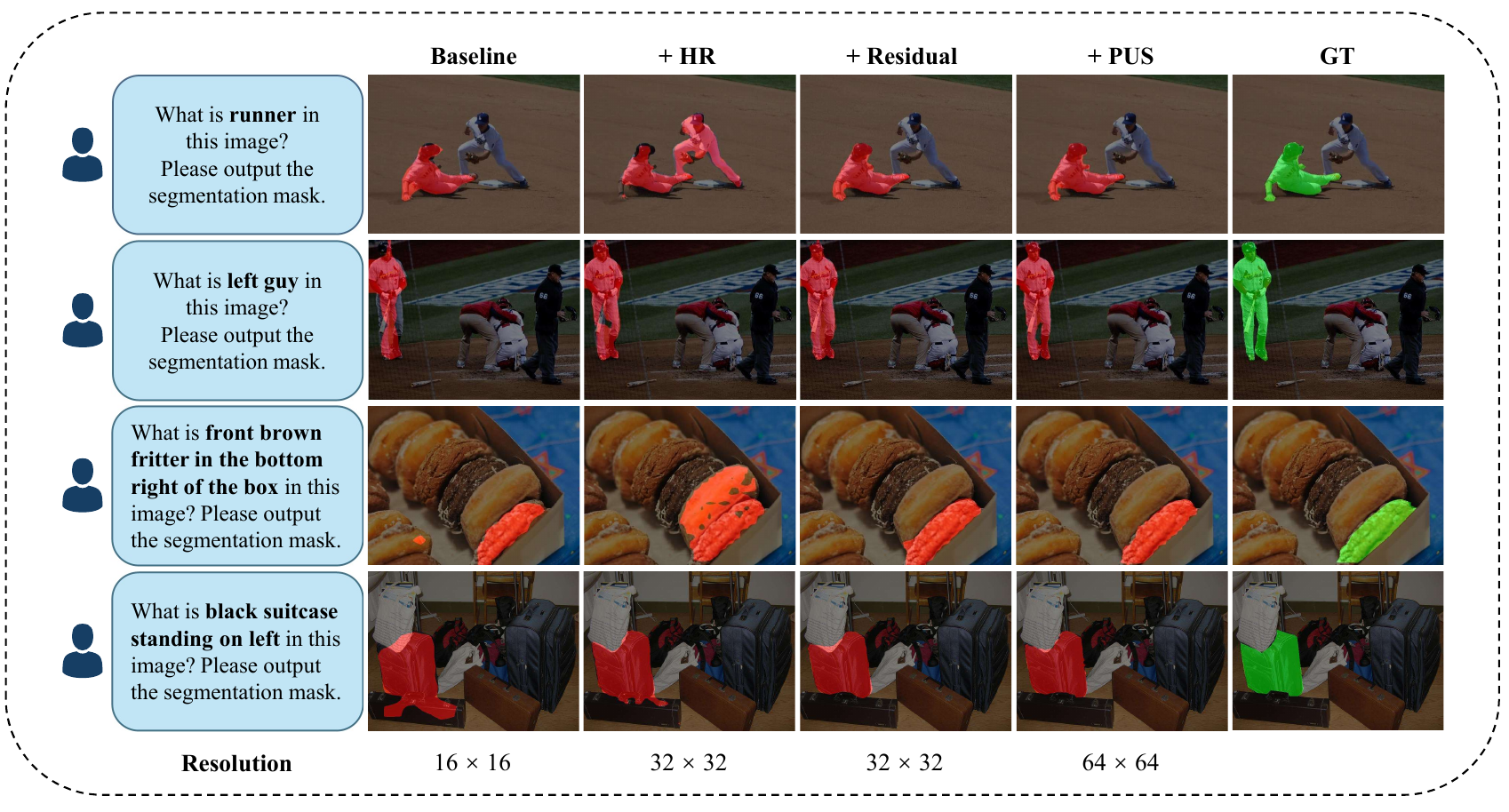}
    \vspace{-2mm}
    \caption{Visualization results on RefCOCO demonstrate the effectiveness of the modules. 
    `HR' stands for Higher Resolution of uncompressed image features, `Residual' refers to the use of residual features from the LLM, and `PUS' represents MLP-with-Pixel-Unshuffle. 
    The bottom row illustrates the resolution of the fused image features and the predicted mask before interpolation to the original image size.}
    \label{fig:refcoco_vis}
\end{figure*}

\begin{table}[!t]
    \centering
    \resizebox{0.7\linewidth}{!}{
    \begin{tabular}{ccc|ccc}
    \toprule
         $\bm{F}_{V_1}^{HQ}$ & RFR & PUS & val & testA & testB \\
         \midrule
         & & & 76.6 & 78.3 & 73.6 \\
         \checkmark & & & 78.7 & 81.1 & 75.2 \\
         \checkmark & \checkmark & & 79.0 & 81.3 & 76.3 \\
         & & \checkmark & 78.5 & 80.3 & 75.1 \\
         \checkmark & \checkmark & \checkmark & \textbf{79.8} & \textbf{82.0} & \textbf{77.1} \\
         \bottomrule
    \end{tabular}
    }
    \caption{Ablation of RFR and Pixel-Unshuffle on RefCOCO.}
    \label{tab:nrpr}
\end{table}

\subsection{Ablation Study}

\subsubsection{Effectiveness of Residual Features Refilling}\label{sec:abl_rfr}

Sec.~\ref{sec:nrpr} introduces the primitive design of constructing higher resolution image features for segmentation. 
We compare the results with different selections of features, as well as the usage of Residual Features Refilling and Pixel-Unshuffle. 
As shown in Tab.~\ref{tab:nrpr}, directly using uncompressed features $\bm{F}_{V_1}^{HQ}$ from the image encoder and Pixel-Shuffle-with-MLP has improvements of 2.1\% on RefCOCO val dataset and 2.8\% on RefCOCO testA dataset, which demonstrates the importance of higher resolution for segmentation. 
Although $\bm{F}_{V_1}^{HQ}$ features are sufficient for coarse target discrimination, they still fall short in capturing fine-grained targets.
% For instance, the RefCOCO testB dataset requires recognizing humans with specific characteristic, position, or action, which could not be easily distinguished from $\bm{F}_{V_1}^{HQ}$. 
For instance, some samples from RefCOCO simultaneously contain objects of the same category or multiple humans in an image, where $\bm{F}_{V_1}^{HQ}$ struggles to distinguish the specific target described in the text. 
Therefore, the integration of residual features from the LLM provides a finer-grained understanding of image features, where the performance on RefCOCO testB dataset increases from 75.2\% to 76.3\% of cIoU. 
Moreover, the Pixel-Unshuffle applied to image features fused with residual features gains a promotion of approximately 2\% on each subset of RefCOCO. 
The combination of RFR and Pixel-Unshuffle achieves the improvements of approximately 0.8\% on each subset, further improving performance for referring segmentation without a specialist mask decoder. 

\subsubsection{Effectiveness of Residual Features Amplifier}\label{sec:abl_rfs}

Comparison of various RFA designs is presented in Tab.~\ref{tab:nrps}. 
The baseline are the vanilla design in Sec.~\ref{sec:nrpr}, which achieves cIoU of 78.9\%, 73.5\%, and 76.3\% on RefCOCO/+/g datasets, respectively. 
We attempt to integrate a pixel-unshuffle process with MLP~(2\textsuperscript{nd} row) for residual features, yet the performance gains are limited promotion with 0.4\% across all tasks. 
Our RFA involves three rounds of pixel-unshuffle with MLP for image features of two resources. 
Results of sharing the same MLP~(3\textsuperscript{rd} row) and using independent MLPs~(5\textsuperscript{th} row) for $\bm{F}_{IMG}$, $\bm{F}_{V_1}$, and $\bm{F}_{V_1}^{HQ}$ both demonstrate enhancement of approximately 0.6\% on average. 
Our design with shared MLP for image features without LLM~($\bm{F}_{V_1}$ and $\bm{F}_{V_1}^{HQ}$) and a different one for LLM-processed image features~($\bm{F}_{IMG}$) achieves an average increase of 1.0\%, thereby underscoring the effectiveness of assigning each MLP to a specific purpose and further validating the efficacy of RFA for higher resolution features. 

\begin{table}[!t]
    \centering
    \resizebox{\linewidth}{!}{
    \begin{tabular}{l|ccc}
    \toprule
          & RefCOCO & RefCOCO+ & RefCOCOg \\
         \midrule
         1. Baseline & 78.9 & 73.5 & 76.3 \\
         2. 1+Part Unshuffle & 79.3 & 73.9 & 76.7 \\
         3. 1+RFA(MLP$\times$1) & 79.4 & \textbf{74.3} & 76.7 \\
         4. 1+RFA(MLP$\times$2) & \textbf{80.0} & \textbf{74.3} & \textbf{77.4} \\    
         5. 1+RFA(MLP$\times$3) & 79.7 & 74.2 & 76.6 \\
         % 6. 1+PUS & 79.6 & 74.2 & 76.8 \\
         % 7. 4+PUS & 80.0 & 74.3 & 77.4 \\
         \bottomrule
    \end{tabular}
    }
    \caption{Ablation study of RFA on Referring Expression Segmentation tasks. The results are the average of cIoU among the subsets. }
    \label{tab:nrps}
\end{table}

\begin{table}[!t]
    \centering
    \resizebox{\linewidth}{!}{
    \begin{tabular}{l|ccc}
    \toprule
          & RefCOCO & RefCOCO+ & RefCOCOg \\
         \midrule
         Causal & 78.4 & 72.7 & 75.5 \\
         +\texttt{[IMG]}$\rightarrow$\texttt{[IMG]} & 79.0 & 73.5 & 75.8 \\
         +\texttt{[IMG]}$\rightarrow$\texttt{[SEG]} & 80.0 & 74.3 & 77.4 \\
         +\texttt{[IMG]}$\rightarrow$\texttt{[TEXT]} & 80.2 & 74.9 & 77.0 \\    
         Bidirectional & 80.2 & 75.1 & 76.9 \\
         \bottomrule
    \end{tabular}
    }
    \caption{Ablation study on the attention masks for segmentation. The causal mask and bidirectional mask denote two widely used attention masks. 
    The ``$\rightarrow$" represents the additional visible regions, \textit{e.g.}, \texttt{[IMG]}$\rightarrow$\texttt{[SEG]} means the \texttt{[SEG]} tokens are visible to the previous \texttt{[IMG]} tokens. }
    \label{tab:ti}
\end{table}

\subsubsection{Analysis of Token Interaction}

We analyze the impact of different attention mask designs in Table \ref{tab:ti}. 
Our baseline~(1\textsuperscript{st} row), the standard causal mask, achieves 78.4\%, 72.7\%, and 75.5\% cIoU on RefCOCO, RefCOCO+, and RefCOCOg, respectively. 
Enabling bidirectional intra-visual interaction~(2\textsuperscript{nd} row) from the causal mask yields only marginal gains of 0.5\% on average.
However, incorporating cross-modal interactions significantly improves performance. 
Allowing \texttt{[IMG]} tokens to investigate the \texttt{[SEG]} tokens~(3\textsuperscript{rd} row) boosts the results to 80.0\%, 74.3\%, and 77.4\%, respectively, which represents a substantial improvement of 1.6\%, 1.6\%, and 1.9\% over the causal baseline, and achieves the best performance on the RefCOCOg dataset. 
Moreover, allowing attention to \texttt{[TEXT]} tokens~(4\textsuperscript{th} row) or using a full Bidirectional mask~(5\textsuperscript{th} row) still has remarkable promotions, yet the intra-visual with \texttt{[IMG]}-to-\texttt{[SEG]} interactions provide almost the same effectiveness with much less renovation from original settings. 

\subsubsection{Analysis of Visualization Results on RefCOCO}
Fig. \ref{fig:refcoco_vis} shows the visualization results that illustrate the impact of various modules on segmentation performance. 
The baseline uses image features directly from the LLM output, while HR uses uncompressed image features from the image encoder. 
The residual column adds residual image features from LLM, and PUS further enhances resolution through MLP-with-Pixel-Unshuffle to uncompressed features fused with residual features.

% As modules are added, segmentation accuracy obviously improves. In the second row, the model increasingly defines the boundaries of the person, and by the end, it even distinguishes the gaps between the legs. In the fourth row, the model focuses more on the target object, effectively eliminating nearby distracting objects.
% However, using HR alone does not always produce the best results. The HR image features lack the segmentation-related information from the LLM, making it harder to distinguish semantically similar objects, as seen in the first and third rows. Thus, combining the LLM's residual features with HR and the up-sampling process yields the best results.

As modules are added, segmentation accuracy progressively improves. The model gradually refines object boundaries and eventually distinguishes fine structures such as gaps between legs~(2\textsuperscript{nd} row). 
It also focuses more accurately on target objects while suppressing surrounding distractions~(4\textsuperscript{th} row). 
However, relying solely on uncompressed features is suboptimal, as they lack segmentation-relevant features from the LLM, leading to confusion between semantically similar objects~(\textit{e.g.} 1\textsuperscript{st} \& 3\textsuperscript{rd} row). 
Combining the LLM’s residual features with HR features and PUS modules achieves the best performance.

\section{Conclusion}
In this paper, we presented \textbf{SELF1E}, to our knowledge, the first MLLM-based segmentation model that operates without a specialist mask decoder while solely relying on a single \texttt{[SEG]} token. 
We introduced RFR and RFA modules to fuse high-resolution image features from the MLLM encoder with segmentation-relevant features from LLM, through interactions between \texttt{[SEG]} and \texttt{[IMG]} tokens. 
The resulting high-resolution and high-quality image features enable accurate pixel-level masks with minimal additional parameters. 
Extensive experiments validate the effectiveness of the proposed SELF1E, achieving state-of-the-art performance across various visual and segmentation tasks. 
% Our approach expands the possibilities of MLLM-based segmentation, providing an effective and lightweight solution while paving the way for the development of more powerful and architecturally simple MLLMs in the future.
\clearpage

\section*{Acknowledgments} 
This work was supported by the National Natural Science Foundation of China (Grant 62472033, 92470203, U23A20314, 61972036), and the Beijing Natural Science Foundation (Grant L242022).

{
    \small
    \bibliographystyle{ieeenat_fullname}
    \bibliography{main}
}

\clearpage
\setcounter{page}{1}
\maketitlesupplementary

\begin{table*}[!t]
    \centering
    \begin{tabular}{c|c|c|c|c}
    \toprule
         Task & Dataset & Samples & SEG-rates & SEG-samples \\
         \midrule
         \multirow{6}{*}{VQA} & VQAv2~\cite{vqav217} & 100k & \multirow{6}{*}{1$\times$} & 100k\\
         & OKVQA~\cite{okvqa19} & 9k & &9k\\
         & TextVQA~\cite{textvqa19} & 35k & & 35k\\
         & VizWiz~\cite{vizwiz18} & 20k & & 20k\\
         & GQA~\cite{gqa19} & 100k & & 100k\\
         & LLaVA-150k~\cite{llava23} & 157k & & 157k\\
         \midrule
         \multirow{3}{*}{Referring Expression Segmentation} & RefCOCO~\cite{refcoco/+14} & 17k & \multirow{3}{*}{20$\times$} & 340k \\
         & RefCOCO+~\cite{refcoco/+14} & 17k & & 340k \\
         & RefCOCOg~\cite{refcoco/+/g16} & 22k && 440k \\
         \midrule
         \multirow{4}{*}{Semantic Segmentation} & ADE20k~\cite{ade20k18} & 20k &\multirow{4}{*}{6$\times$} & 120k \\
         & COCOStuff~\cite{cocostuff18} & 30k & & 180k \\
         & Pascal-Part~\cite{pascal_part14} & 4k & & 24k \\
         & LVIS-PACO~\cite{paco23} & 30k & & 180k \\
         \midrule
         Reasoning Segmentation & ReasonSeg~\cite{lisa24} & 239 & 6$\times$ & 1.4k \\
         \midrule
         Overall & & 561k&& 2.4M\\
         \bottomrule
    \end{tabular}
    \caption{Details of the multiple datasets for training. SEG-rates represent the magnification of the dataset samples for the training of SELF1E-SEG version. }
    \label{tab:datasets}
\end{table*}

\begin{table*}[!t]
    \centering
    \begin{tabular}{c|ccccccccc}
    \toprule
         Methods & VQAv2 & OKVQA & VizWiz & GQA & TextVQA & POPE & MMB-en & MMB-cn & MME \\
         \midrule
         InternVL3-2B~\cite{internvl325} & 80.1 & 56.1 & 56.6 & 60.7 & 77.0 & 89.6 & 81.1 & 78.4 & 2221.2\\
         InternVL3-8B~\cite{internvl325} & 81.8& 61.9& 63.4& 63.2& 78.9& 91.1& 83.4& 82.2&2415.4\\
         SELF1E-2B  & 77.7& 48.5& 67.8& 61.5& 71.7& 89.4& 69.6& 66.2& 2014.6\\
         SELF1E-8B  & 80.2& 54.0& 69.9& 64.1& 72.3& 91.1& 74.6& 72.0& 2265.1\\
         \bottomrule
    \end{tabular}
    \caption{Comparison of the VQA performance of our approach with their original base MLLMs. }
    \label{tab:vqaexp}
\end{table*}

\begin{table*}[!t]
    \centering
    \resizebox{0.9\linewidth}{!}{
        \begin{tabular}{r|c|ccc|ccc|cc}
        \toprule 
        \multirow{2}{*}{ Method } & \multirow{2}{*}{MLLM} & \multicolumn{3}{c|}{ RefCOCO } & \multicolumn{3}{c|}{ RefCOCO+ } & \multicolumn{2}{c}{ RefCOCOg }\\
        & & val & testA & testB & val & testA & testB & val & test \\
        \midrule
        
        \multirow{3}{*}{SELF1E-2B} & \cellcolor{gray!30} InternVL3-2B & \cellcolor{gray!30} 80.2 & \cellcolor{gray!30} 82.1 & \cellcolor{gray!30} 77.6 & \cellcolor{gray!30} 74.6 & \cellcolor{gray!30} 79.1 & \cellcolor{gray!30} 69.2 & \cellcolor{gray!30} 77.0 & \cellcolor{gray!30} 77.8 \\
        & InternVL2-2B & 77.7 & 80.6 & 74.6 & 71.5 & 76.7 & 66.9 & 74.3 & 74.7 \\
        & InternVL2.5-2B & 80.1 & 82.2 & 78.0 & 74.7 & 78.7 & 69.8 & 76.5& 77.6 \\
        \midrule
        
        \multirow{3}{*}{SELF1E-SEG-2B} & \cellcolor{gray!30} InternVL3-2B & \cellcolor{gray!30} 84.3 & \cellcolor{gray!30} 85.4 & \cellcolor{gray!30} 82.3 & \cellcolor{gray!30} 78.9 & \cellcolor{gray!30} 83.5 & \cellcolor{gray!30} 75.1 & \cellcolor{gray!30} 80.4 & \cellcolor{gray!30} 81.0 \\
        & InternVL2-2B & 83.5 & 85.5 & 81.4 & 77.7 & 82.0 & 72.9 & 79.5 & 80.1 \\
        & InternVL2.5-2B & 85.2 & 86.7 & 83.5 & 79.9 & 83.4 & 75.2 & 81.0 & 82.3 \\
        % SELF1E-8B & $\checkmark$ &  $\checkmark$ & 82.5 & 83.8 & 80.1 & 77.6 & 81.6 & 73.7 & 79.1 & 79.8 \\
        % SELF1E-SEG-8B & $\checkmark$ &  $\checkmark$ & 84.7 & \textbf{86.2} & \underline{83.4} & \textbf{80.2} & \textbf{84.2} & \textbf{77.0} & \textbf{82.1} & \textbf{82.8} \\
        \bottomrule
        \end{tabular}
    }
    \caption{Comparison of results with different MLLMs as base model on the Referring Expression Segmentation benchmarks (RefCOCO/+/g). }
    \label{tab:mllmab}
\end{table*}

\section{Limitations}

Our method achieves competitive results on various segmentation tasks, yet the limitations still exist. 
The token interactions among the \texttt{[IMG]} tokens and \texttt{[SEG]} token enhance the spatial precision of features, yet the redesigned attention mask becomes an obstacle for autoregressive inference and multi-round reasoning. 
We could only predefine the text templates or separate the process of text inference and segmentation as compromise. 
% Moreover, since our model predicts a $64 \times 64$ mask that is subsequently interpolated to the original resolution—which may reach several hundreds or even thousands of pixels—the interpolation process can introduce inaccuracies.
% In particular, objects with highly curved boundaries or thin structures may not be reconstructed with perfect accuracy after upsampling.
% While this limitation is inherent to many segmentation methods that rely on low-resolution mask predictions, it remains a potential direction for further improvement in future work.
Besides, the original VQA capabilities  of MLLMs are not fully preserved according to Sec.~\ref{sec:vqaexp}. 
The enhanced localization and grounding capabilities from segmentation samples conflict with OCR-oriented and more complex knowledge reasoning scenarios, presenting a promising direction for future work toward better balance.

\section{Details about training datasets}

We utilize a broad collection of vision–language and pixel-level segmentation datasets to train both the base version of SELF1E and the segmentation-enhanced SELF1E-SEG. 
The details are shown in Tab.~\ref{tab:datasets}. 
The VQA component is constructed from six datasets, where VQAv2~\cite{vqav217} provides large-scale human-annotated question–answer pairs for general vision understanding, LLaVA-150k~\cite{llava23} offers high-quality multimodal conversational annotations from GPT-4, and OKVQA~\cite{okvqa19}, TextVQA~\cite{textvqa19}, VizWiz~\cite{vizwiz18}, and GQA~\cite{gqa19} further contribute knowledge-based, text-centric, low-quality-image, and compositional reasoning supervision. 
These datasets are incorporated without magnification (1$\times$) for SEG version, totaling 421k samples. 
For language-guided referring expression segmentation, we adopt the RefCOCO, RefCOCO+, and RefCOCOg datasets~\cite{refcoco/+/g16,refcoco/+14}, which feature object-level referring expressions with increasing linguistic complexity. 
These datasets are expanded by a 20$\times$ SEG-rate, providing 1.12M effective samples for SELF1E-SEG. 
To strengthen dense pixel-level perception, we employ ADE20K~\cite{ade20k18} that covers a broad spectrum of indoor/outdoor scenes with fine-grained masks, along with COCO-Stuff~\cite{cocostuff18} and Pascal-Part~\cite{pascal_part14} for diverse semantic regions and part-level annotations, and LVIS-PACO~\cite{paco23}, which supplies long-tailed, instance-rich perceptual concepts. 
Each dataset is magnified 6$\times$, yielding 504k samples for SEG version. 
Finally, ReasonSeg~\cite{lisa24} is included to support more complex reasoning-driven segmentation, where its limited 239 samples are expanded 6× into approximately 1.4k effective instances. 
Overall, our training corpus comprises roughly 561k samples for the base version SELF1E and around 2.4M magnified samples for SELF1E-SEG. 

\section{Additional Experiment Results}

\subsection{Experiment results of VQA}\label{sec:vqaexp}

Across 2B and 8B model scales, SELF1E exhibits a consistent performance pattern when compared with the corresponding InternVL3 baselines, as illustrated in Tab.~\ref{tab:vqaexp}. 
On generic benchmarks such as VizWiz, GQA, and VQAv2, SELF1E consistently achieves similar results as InternVL3, especially with gains of 11.2\% on 2B and 6.5\% on 8B on VizWiz and moderate improvements on GQA. 
These results suggest that introducing segmentation-aware visual supervision could retain the original generic understanding ability of the images. 
By contrast, SELF1E shows lower performance on OKVQA and TextVQA at both scales. 
Since these benchmarks heavily depend on external knowledge grounding (OKVQA) or OCR-oriented textual reasoning (TextVQA), the performance gap indicates that segmentation-focused training, provides limited improvement in text-heavy or knowledge-intensive settings even with specific training data.
A similar trend is shown on instruction-oriented multimodal benchmarks (MMB-en/cn, MME), where SELF1E trails InternVL3 regardless of scale. Most of the performance reduction is on the OCR-oriented sub-tasks and more complex knowledge sub-tasks. Nevertheless, SELF1E maintains competitive POPE scores across scales, matching or approaching InternVL3, demonstrating that stronger spatial grounding introduced by segmentation has a limited negative influence on hallucination.

Overall, the unified comparison across both 2B and 8B models demonstrates that the strengths of SELF1E lie primarily in perception robustness and grounding-oriented reasoning, enabled by segmentation-enhanced visual modeling, whereas performance trade-offs emerge on OCR-heavy and knowledge-driven benchmarks. This consistent pattern across scales highlights the complementary nature of segmentation-aware learning within MLLMs and reveals clear future directions for balancing visual grounding with textual and knowledge-centric capabilities.

\subsection{Experiment results with other MLLMs}

We conduct several experiments of SELF1E based on different versions of MLLMs in Tab.~\ref{tab:mllmab}, including InternVL2-2B and InternVL2.5-2B, on which some of the previous methods applied. 
The results show that even with previous versions of InternVL, our approach still achieves competitive performance. 
Using InternVL2.5-2B as the base model attains similar results on the standard version of SELF1E, while having approximately 1\% of advantage on the SELF1E-SEG version. 
In summary, the state-of-the-art performance of our SELF1E does not heavily rely on the upgrade of MLLM, where it still has advanced performance with earlier versions of InternVL. 

\begin{table}[!t]
    \centering
    \resizebox{\linewidth}{!}{
    \begin{tabular}{l|ccc}
    \toprule
          & RefCOCO & RefCOCO+ & RefCOCOg \\
         \midrule
         Baseline& 76.2& 72.3& 74.5\\
         Scanning& 78.3& 73.3& 76.3\\
         Self-Replication& 78.9& 73.5& 76.3\\
         % Interpolation & 79.3 & 74.1 & 76.5 \\
         \bottomrule
    \end{tabular}
    }
    \caption{Ablation study on the operations for retaining the resolution. }
    \label{tab:sr}
\end{table}

\subsection{Ablation Study of Retaining Resolution}

The RFR operation in Sec.~\ref{sec:nrpr} requires uncompressed image features for retaining the original resolution of image features from the encoder. 
Thus, we design an experiment to measure the effectiveness of different strategies. 
As shown in Tab.~\ref{tab:sr}, we compare the self-replication strategy with the scanning strategy. 
To be specific, the original pixel-shuffle process has the same stride value as the factor, so that different groups of features are not overlapped. 
The scanning strategy set the stride as 1, which preserves the original resolution. 
However, the results with scanning strategy, although higher than the baseline without any strategy, are still slightly lower than self-replication 0.6\% on RefCOCO and 0.2\% on RefCOCO+. 
The single pixel features from the scanning strategy are generated from $\alpha$ pixels from pre-compressed image features, while those from the self-replication strategy only correspond to the same pixel that preserves more precise spatial details. 

\subsection{Efficiency Comparison}

To clarify when decoder-free is preferable, we report inference efficiency in Tab.~\ref{tab:speed_comparison} based on a single NVIDIA RTX4090.
SELF1E achieves the fastest inference, outperforming LISA with specialist segmentation decoder and UFO with multi-token prediction.
Even without customization for higher efficiency, SELF1E is still more memory-efficient than LISA and significantly faster than UFO, as it eliminates computational efforts on auxiliary mask decoders and multi-token decoding.

\begin{table}[h]
    \centering
    \caption{Efficiency comparison among LISA, UFO, and SELF1E.}
    \label{tab:speed_comparison}
    \small
    \begin{tabular}{lccc}
        \hline
        Method & Inference (ms) & FPS & Memory (GB) \\
        \hline
        LISA-7B  & 250.0 & 4.00 & 19.2 \\
        UFO-7B  & 961.5  & 1.04 & 14.4 \\
        \textbf{SELF1E-8B} & \textbf{105.0} & \textbf{9.52} & 17.7 \\
        \hline
    \end{tabular}
    % \label{tab:compute_effi}
\end{table}

\subsection{More Analysis on VQA}

\begin{figure}[h]
    \centering
    \includegraphics[width=\linewidth]{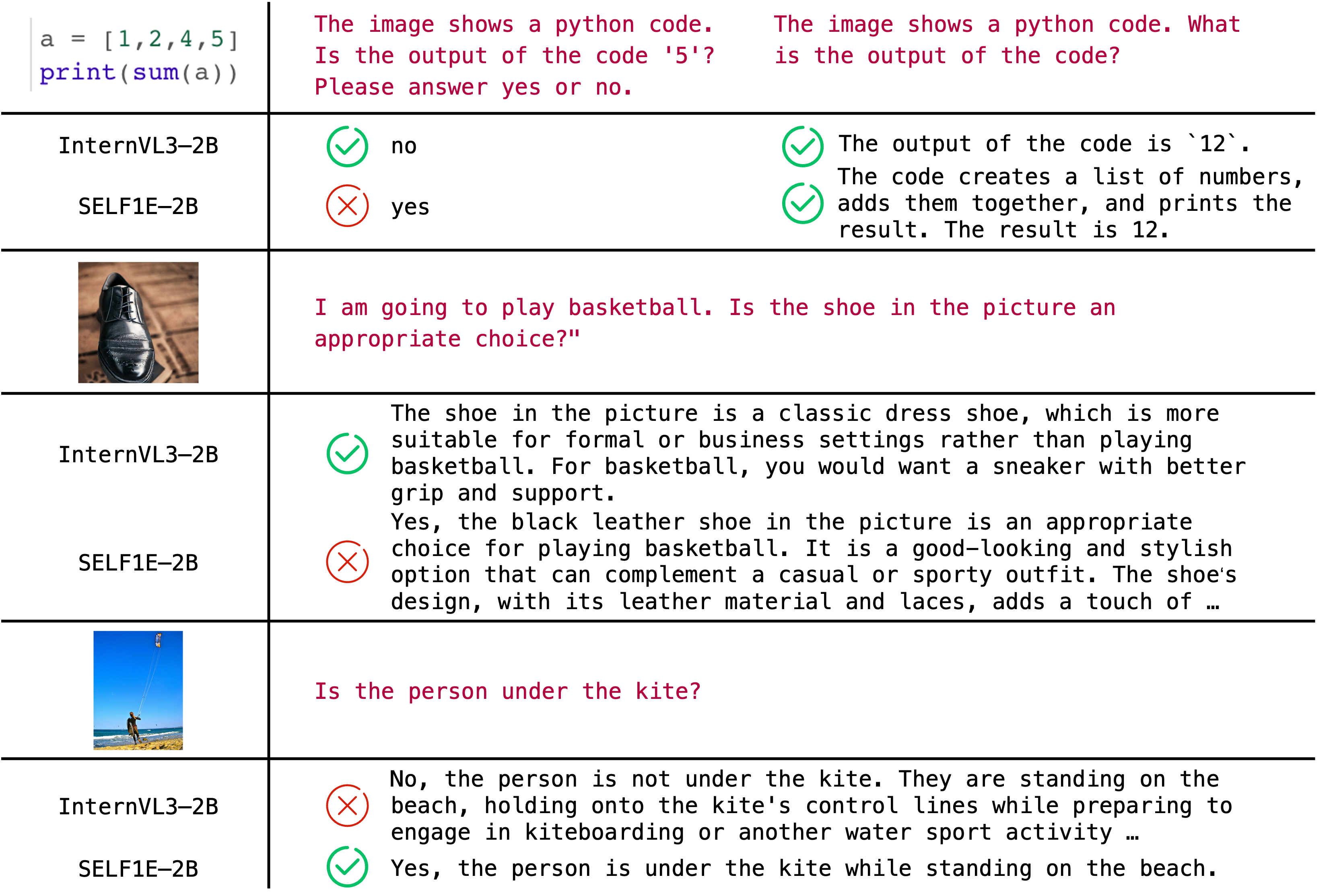}
    \caption{Representative VQA results comparison. }
    \label{fig:vqasamples}
\end{figure}

\begin{table*}[h]
\centering
\caption{MME Benchmark Performance Comparison}
\label{tab:mme_comparison}
% \resizebox{\linewidth}{!}{%
\begin{tabular}{llcccccccccc}
\hline
 & \textbf{Method} & \textbf{Exist.} & \textbf{Count} & \textbf{Pos.} & \textbf{Color} & \textbf{Post.} & \textbf{Celeb.} & \textbf{Scene} & \textbf{Landm.} & \textbf{Artw.} & \textbf{OCR} \\ 
 \hline
\multirow{2}{*}{\rotatebox[origin=c]{90}{\textbf{Perp.}}} & SELF1E & 190.00 & 140.00 & 153.33 & 175.00 & 164.29 & 159.41 & 160.25 & 156.50 & 140.50 & 102.50 \\
 & InternVL3-2B & 195.00 & 165.00 & 136.67 & 170.00 & 157.48 & 155.00 & 156.25 & 163.75 & 156.75 & 155.00 \\ 
 \hline
\hline
 & \textbf{Method} & \multicolumn{2}{c}{\textbf{Comm.}} & \multicolumn{2}{c}{\textbf{Num.}} & \multicolumn{2}{c}{\textbf{Trans.}} & \multicolumn{2}{c}{\textbf{Code}} & \multicolumn{2}{c}{\textbf{Total Score}} \\ 
 \hline
\multirow{2}{*}{\rotatebox[origin=c]{90}{\textbf{Cog.}}} & SELF1E & \multicolumn{2}{c}{97.86} & \multicolumn{2}{c}{100.00} & \multicolumn{2}{c}{162.50} & \multicolumn{2}{c}{112.50} & \multicolumn{2}{c}{\textbf{2014.64}} \\
 & InternVL3-2B & \multicolumn{2}{c}{115.71} & \multicolumn{2}{c}{105.00} & \multicolumn{2}{c}{185.00} & \multicolumn{2}{c}{147.50} & \multicolumn{2}{c}{\textbf{2164.11}} \\ 
 \hline
\end{tabular}%
% }
\end{table*}
As shown in Tab.~\ref{tab:mme_comparison}, incorporating pixel-level supervision leads to moderate degradation on knowledge-intensive and OCR-related tasks (e.g., Artworks, OCR, Commonsense, Code), while improving spatial understanding (Position).
This indicates that segmentation supervision biases the model toward spatial grounding at some cost to abstract reasoning.
Representative VQA examples in Fig.~\ref{fig:vqasamples} further illustrate this effect, where spatial and positional queries improve while format-sensitive or multi-step reasoning may degrade.
Overall, SELF1E preserves general VLM capability reasonably well while making an explicit and transparent trade-off to enable high-quality segmentation.

\section{Prompt Settings for Segmentation Tasks}

% \subsection{Prompts}
\subsection{Training}
Our templates inherit the design principle of LISA~\cite{lisa24}. Different dataset types use different prompt templates during training to align with their annotation styles.

For semantic segmentation task and vanilla referring expression segmentation task, we refer to the category name or object description simply as \textit{text} for convenience:

We define a short question list:
\begin{itemize}
    \item \texttt{Can you segment the \{\textit{text}\} in this image?}
    \item \texttt{Please segment the \{\textit{text}\} in this image.}
    \item \texttt{What is \{\textit{text}\} in this image? Please respond with segmentation mask.}
    \item \texttt{What is \{\textit{text}\} in this image? Please output segmentation mask.}
\end{itemize}

and a answer list:
\begin{itemize}
    \item \texttt{It is [SEG].}
    \item \texttt{Sure, [SEG].}
    \item \texttt{Sure, it is [SEG].}
    \item \texttt{Sure, the segmentation result is [SEG].}
    \item \texttt{[SEG].}
\end{itemize}

The full template can be represent as:

\texttt{%
\textbf{USER}: <IMG> \{a random choice from short question list\}
}

\texttt{%
\textbf{ASSISTANT}: \{a random choice from answer list\}
}

For reasoning segmentation task, the query expands into a longer, implicit instruction:

We use a long question list as below when the instruction is a full sentence; otherwise, we apply the short question list:
\begin{itemize}
    \item \texttt{\{\textit{instruction}\} Please respond with segmentation mask.}
    \item \texttt{\{\textit{instruction}\} Please output segmentation mask.}
\end{itemize}

The full template can be represent as:

\texttt{%
\textbf{USER}: <IMG> \{a random choice from long or short question list\}
}

\texttt{%
\textbf{ASSISTANT}: \{a random choice from answer list\}
}

% Additionally, for explanatory queries, we employ a specialized explanatory question list:
% \begin{itemize}
%     \item \texttt{\{\textit{instruction}\} Please output segmentation mask and explain why.}
%     \item \texttt{\{\textit{instruction}\} Please output segmentation mask and explain the reason.}
%     \item \texttt{\{\textit{instruction}\} Please output segmentation mask and give some explanation.}
% \end{itemize}

% The full template changes to:

% \texttt{%
% \textbf{USER}: <IMG> \{a random choice from explanatory question list\}
% }

% \texttt{%
% \textbf{ASSISTANT}: \{a specific answer explaining the question\}
% }

\subsection{Validation}
Our validation prompts follow two instruction formats, depending on whether the dataset provides object names or full-sentence instructions. Below, we provide the exact input–output templates used during validation.

For giving a specific object or description (\textit{i.e.} RES and OVS datasets):

\texttt{%
\textbf{USER}: <IMG> What is \{\textit{object's name or description}\} in this image? Please output segmentation mask.
}

\texttt{%
\textbf{ASSISTANT}: [SEG].
}

For giving a full sentence as instruction (\textit{i.e.} ReasonSeg datasets):

\texttt{%
\textbf{USER}: <IMG> \{\textit{Instruction}\} Please output segmentation mask.
}

\texttt{%
\textbf{ASSISTANT}: [SEG].
}

\section{Additional Visualization Results}

\subsection{Reasoning Segmentation}
\begin{figure*}[!t]
    \centering
    \includegraphics[height=0.95\textheight]{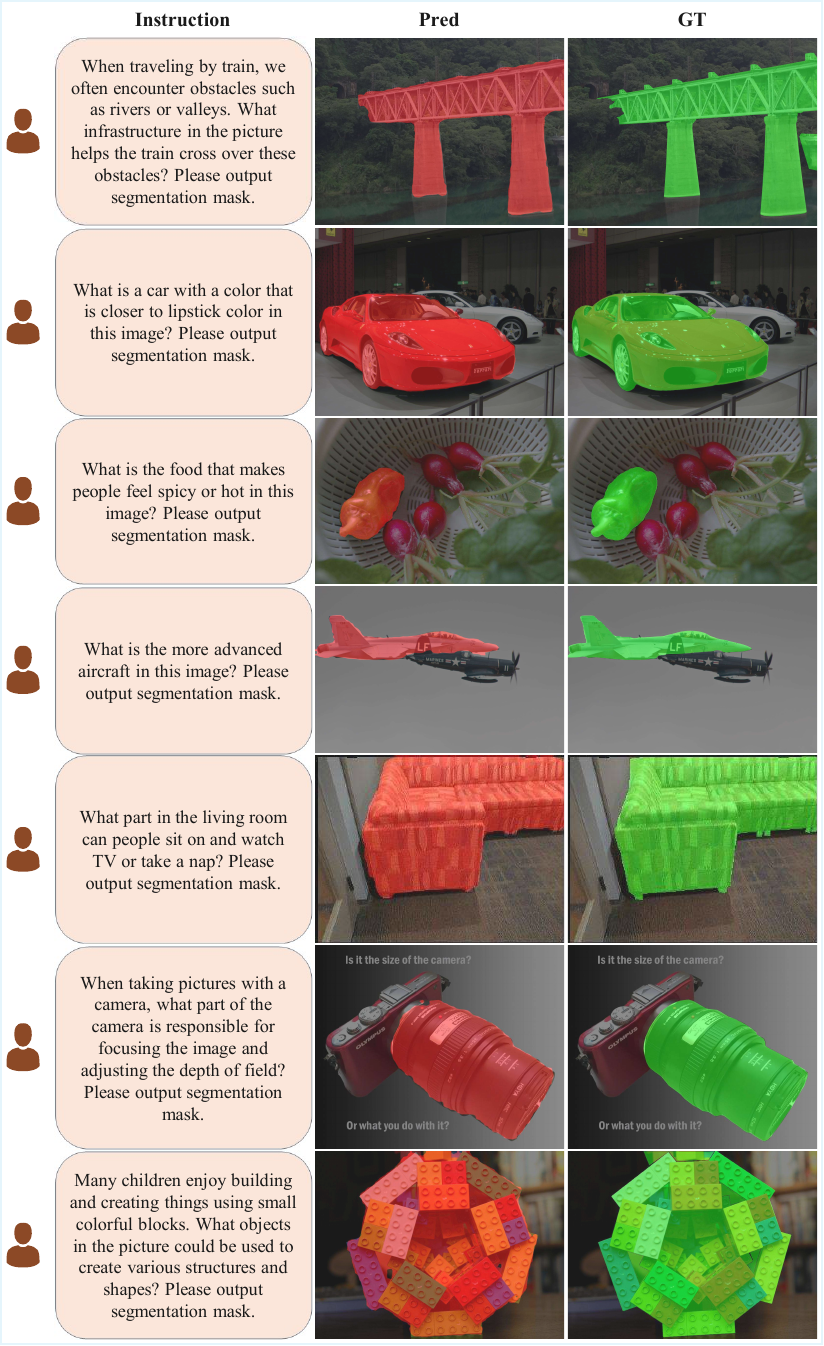}
    \vspace{-2mm}
    \caption{Visualization results on ReasonSeg demonstrate outstanding reasoning ability of SELF1E. “Pred” denotes the predictions from SELF1E, “GT” denotes the ground-truth masks.}
    \label{fig:reason_vis}
\end{figure*}

Reasoning segmentation requires model to infer the correct target object from implicit instructions, rather than relying on explicit object names. As shown in Fig.~\ref{fig:reason_vis}, SELF1E demonstrates strong capability in interpreting complex linguistic instructions and localizing the correct regions with high spatial precision. 
Although our architectural modifications primarily focus on visual features, the LLM’s reasoning ability remains unaffected, retaining its powerful linguistic inference capacity.
Furthermore, the increase in the mask's native resolution provides more detailed structural cues, enabling the model to better capture fine object boundaries. Overall, these results confirm that SELF1E maintains strong reasoning capabilities while benefiting from enhanced visual precision, leading to accurate segmentation under complex reasoning instructions.

\subsection{Open-Vocabulary Segmentation}
\begin{figure*}[!t]
    \centering
    \includegraphics[height=0.95\textheight]{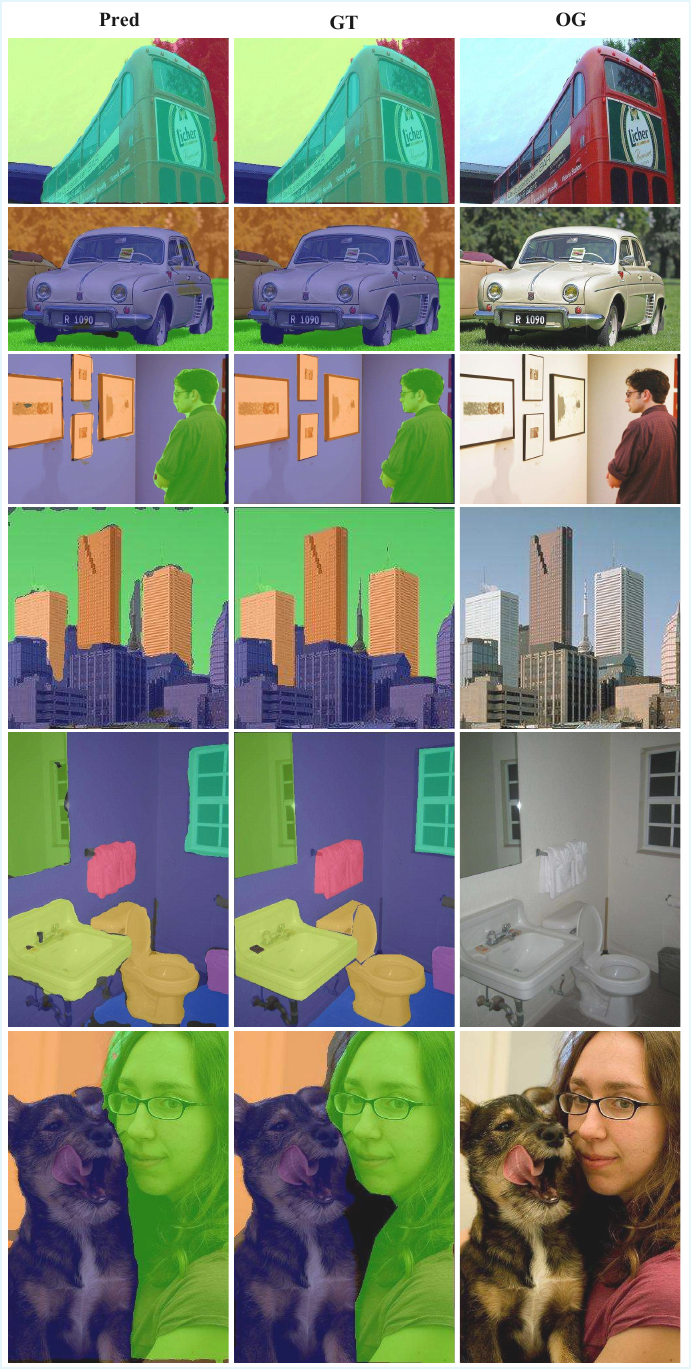}
    \vspace{-2mm}
    \caption{Visualization results on open-vocabulary segmentation. “OG” refers to the original image without overlays.}
    \label{fig:ovs_vis}
\end{figure*}

Fig.~\ref{fig:ovs_vis} presents the visualization results for open-vocabulary segmentation. It is important to note that masks with the same color across different images do not represent the same category; they are merely used to distinguish different objects within a single image. The main challenges in OVS lie in segmenting all instances of a category within an image and accurately distinguishing objects at boundaries, especially when multiple semantically similar objects are present.
From our visualizations, SELF1E performs robustly even in images containing many categories. The model demonstrates precise classification, effectively distinguishing semantically similar objects and accurately capturing object boundaries. These results highlight the model’s strong generalization and fine-grained segmentation capabilities in complex, multi-category scenarios.

\subsection{Token Interaction}
\begin{figure*}[!t]
    \centering
    \includegraphics[height=0.90\textheight]{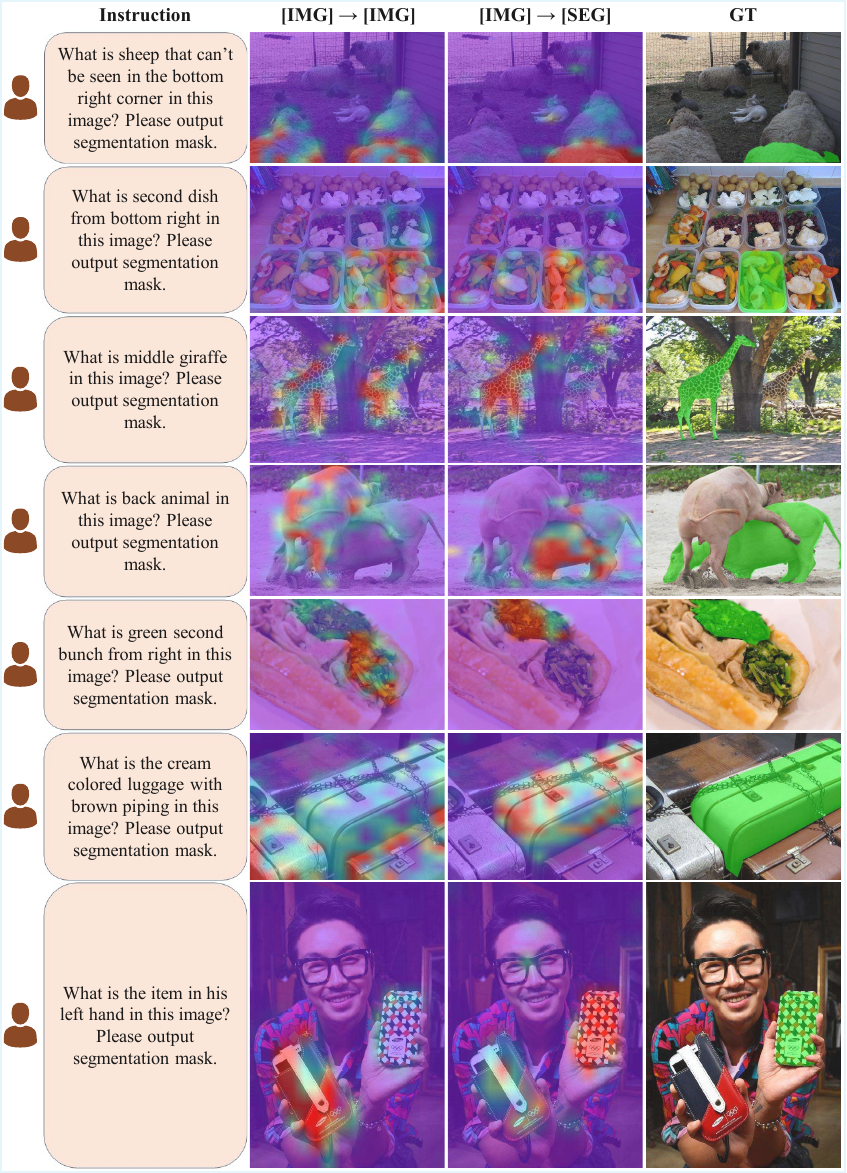}
    \vspace{-2mm}
    \caption{Visualization results show the attention maps of \texttt{[SEG]} to \texttt{[IMG]} tokens under different attention-mask designs. “\texttt{[IMG]$\rightarrow$[IMG]}” indicates that all image tokens use a bidirectional attention mask, while all other tokens follow a causal mask. “\texttt{[IMG]$\rightarrow$[SEG]}” means that, in addition to the bidirectional mask among image tokens, all image tokens are also allowed to interact with the \texttt{[SEG]} token.}
    \label{fig:attn_vis}
\end{figure*}

We visualize the effects of different token interaction strategies on the RES task. Figure~\ref{fig:attn_vis} presents the attention maps from the \texttt{[SEG]} token to \texttt{[IMG]} tokens. When only \texttt{[IMG]} to \texttt{[IMG]} attention is applied, the model is unable to access segmentation-related semantic cues from the \texttt{[SEG]} token. As a consequence, the attention maps sometimes fail to distinguish objects with similar semantics, particularly when the instruction specifies one target among multiple semantically related objects. This limitation becomes even more pronounced for location-dependent queries, where the model may incorrectly allocate high attention to a semantically similar but spatially incorrect object. These observations demonstrate the effectiveness of our \texttt{[IMG]$\rightarrow$[SEG]} token interaction strategy, which substantially alleviates the issues discussed above.

\end{document}